\documentclass[twocolumn]{svjour3}          

\usepackage{times}
\usepackage{epsfig}
\usepackage{amsmath}
\usepackage{amssymb}
\usepackage{array}
\usepackage{subfigure}
\usepackage{multirow}
\usepackage{placeins}
\usepackage{rotating}
\usepackage{varwidth}
\usepackage[usenames,dvipsnames]{xcolor}
\usepackage{footmisc}

\usepackage{latexsym}
\usepackage{natbib}

\usepackage[english]{babel}

\usepackage{url}
\usepackage{xspace}

\newif\ifrevfinal
\revfinalfalse
\def\rev[#1][#2]{\ifrevfinal #2 \else {\color{blue} \sout{#1}} {\bf \color{red} #2} \fi}
\usepackage{color}
\usepackage[normalem]{ulem}

\def\etal{et al.\ }

\def\R{{\rm I\!R}}                            	
\def\<{\langle}
\def\>{\rangle}

\def\be{\begin{equation}}
\def\ee{\end{equation}}
\def\bea{\begin{eqnarray}}
\def\eea{\end{eqnarray}}
\def\fig#1{Figure~\ref{fig:#1}}
\def\tab#1{Table~\ref{tab:#1}}
\def\sect#1{Section~\ref{sec:#1}}

\makeatletter
\gdef\SetFigFont#1#2#3{%
  \reset@font\fontsize{10}{12pt}%
  \selectfont%
}
\let\eqnarray@=\eqnarray \let\endeqnarray@=\endeqnarray
\def\eqnarray{\bgroup\arraycolsep=2pt\eqnarray@}
\def\endeqnarray{\endeqnarray@\egroup}
\makeatother

\def\em{\slshape}

\usepackage{times,amsmath,amssymb,amsbsy,xspace}
\makeatletter
\DeclareRobustCommand\onedot{\futurelet\@let@token\@onedot}
\def\@onedot{\ifx\@let@token.\else.\null\fi\xspace}

\def\eg{\emph{e.g}\onedot} 
\def\ie{\emph{i.e}\onedot} 
 
\def\etc{\emph{etc}\onedot}  
\def\wrt{w.r.t\onedot} 
\def\etal{\emph{et al}\onedot}

\newcounter{iictr}
\def\@iia[#1]{\setcounter{iictr}{#1}\@iib}
\def\@iib{\iii[\roman{iictr}]\xspace}
\def\ii{\@ifnextchar[{\@iia}{\stepcounter{iictr}\@iib}}
\def\iii[#1]{({\em#1\/})}
\makeatother

\def\trecvid{\textsc{TRECVID}\xspace}

\DeclareMathOperator{\maximize}{maximize}

\journalname{International Journal of Computer Vision}

\begin{document}

\title{A robust and efficient video representation for action recognition}


\author{
Heng Wang
\and Dan Oneata
\and Jakob Verbeek
\and Cordelia Schmid
}
\institute{H. Wang is currently with Amazon Research Seattle, but carried out the work described here while being affiliated with INRIA.
All other authors are at INRIA, Grenoble, France.\\
              \email{fistname.lastname@inria.fr}           
}

\date{Received: date / Accepted: date}

\maketitle

\begin{abstract}

This paper introduces a state-of-the-art video representation and
applies it to efficient action recognition and detection. We first
propose to improve the popular dense trajectory features by explicit
camera motion estimation. More specifically, we extract feature point
matches between frames using SURF descriptors and dense optical flow.
The matches are used to estimate a homography with RANSAC. To improve
the robustness of homography estimation, a human detector is employed to
remove outlier matches from the human body as human motion is not constrained
by the camera. Trajectories consistent with the homography are considered
as due to camera motion, and thus removed. We also use the homography to
cancel out camera motion from the optical flow. This results in significant
improvement on motion-based HOF and MBH descriptors.
We further explore the recent Fisher vector as an alternative feature encoding
approach to the standard bag-of-words histogram, and consider different ways to include spatial layout information in these encodings.
We present a large and
varied set of evaluations, considering (i) classification of short basic actions
on six datasets, (ii) localization of such actions in feature-length movies,
and (iii) large-scale recognition of complex events. We find that our improved
trajectory features significantly outperform previous dense trajectories,
and  that Fisher vectors are superior to bag-of-words encodings for  video recognition tasks.
In all three tasks, we show substantial improvements over the state-of-the-art results.


\end{abstract}

\section{Introduction}
\label{sec:introduction}

\begin{figure*}
\centering
 \newcommand{\img}[1]{\includegraphics[width=0.245\linewidth,height=0.14\linewidth]{figures_compare_flow_#1.jpg}}
 \begin{tabular}{c@{\hspace{0.1em}}c@{\hspace{0.1em}}c@{\hspace{0.1em}}c}
  \img{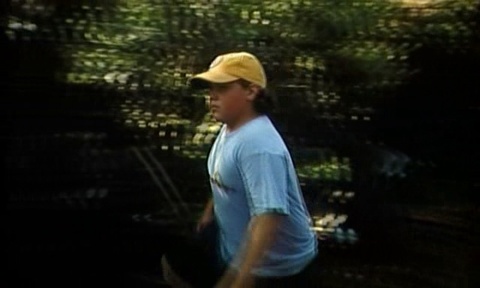} & \img{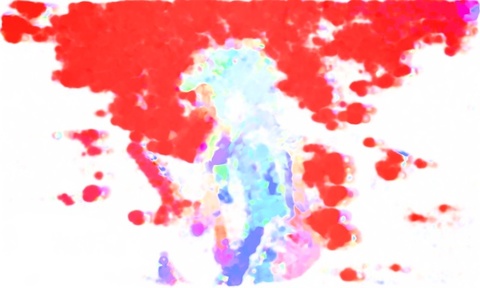} & \img{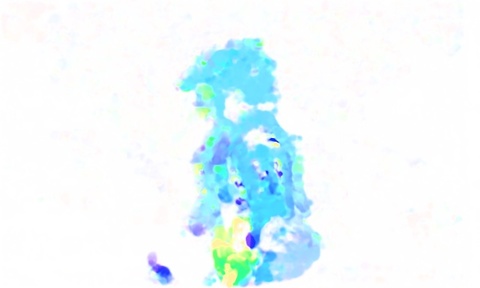} & \img{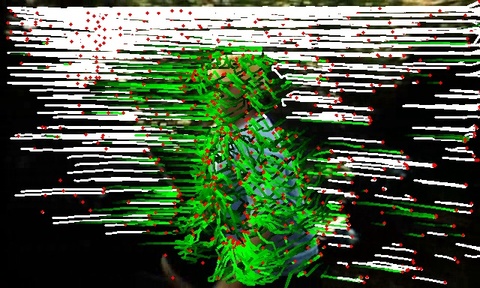} \vspace{-0.25em} \\
  \img{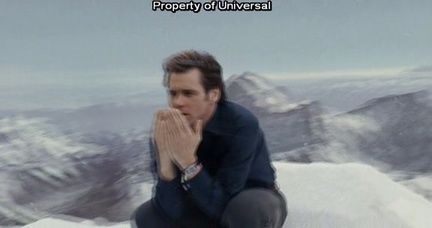} & \img{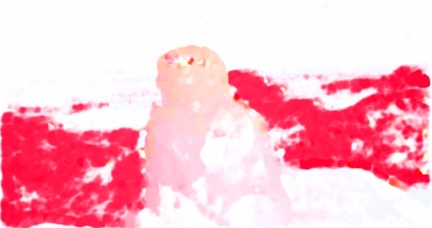} & \img{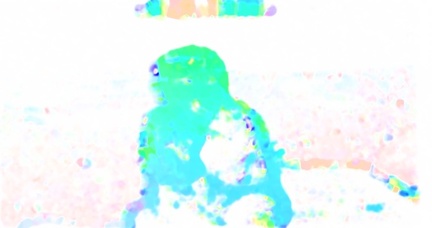} & \img{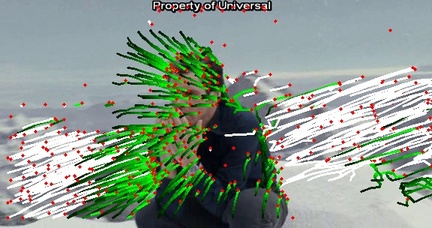} \vspace{-0.25em} \\
 \end{tabular}
 \vspace{0.5em}
\caption{First column: images of two consecutive frames overlaid; second column:
optical flow \citep{Farneback2003} between the two frames; third column: optical
flow after removing camera motion; last column: trajectories removed due to
camera motion in white.}
\label{fig:compare_flow}
\end{figure*}

Action and event recognition have been an active research topic for over
three decades due to their wide applications in video surveillance, human
computer interaction, video retrieval, \etc. Research in this area used
to focus on simple datasets collected from controlled experimental settings,
\eg, the KTH \citep{schuldt2004} and Weizmann \citep{Gorelick2007} datasets.
Due to the increasing amount of video data available from both internet
repositories and personal collections, there is a strong demand for understanding
the content of real world complex video data. As a result, the attention of
the research community has shifted to more realistic datasets such as the
Hollywood2 dataset \citep{Marszalek2009} or the \trecvid Multimedia Event
Detection (MED) dataset \citep{over2012}.

The diversity of realistic video data has resulted in different challenges for action
and event recognition. First, there is tremendous intra-class variation caused
by factors such as the style and duration of the performed action. In addition
to background clutter and occlusions that are also encountered in image-based
recognition, we are confronted with variability due to camera motion, and motion
clutter caused by moving background objects. Challenges can also come from
the low quality of video data, such as noise due to the sensor, camera jitter,
various video decoding artifacts, \etc. Finally, recognition in video also
poses computational challenges due to the sheer amount of data that needs to
be processed, particularly so for large-scale datasets such as the 2014 edition of the \trecvid MED dataset which contains over 8,000 hours of video.

Local space-time features \citep{Dollar2005,Laptev2005} have been shown to be
advantageous in handling such datasets, as they allow to directly build
efficient video representations without non-trivial pre-processing steps,
such as object tracking or motion segmentation. Once local features are
extracted, often methods similar to those used for object recognition are
employed. Typically, local features are quantized, and their overall distribution
in a video is represented with bag-of-words histograms,
see, \eg, \citep{Kuehne2011,Wang2009} for recent evaluation studies.

The success of local space-time features leads to a trend of generalizing
classical descriptors from image to video, \eg, 3D-SIFT \citep{Scovanner2007},
extended SURF \citep{Willems2008}, HOG3D \citep{Klaser2008}, and local trinary
patterns \citep{Yeffet2009}. Among the local space-time features, dense
trajectories \citep{Wang2013} have been shown to perform the best on a variety of
datasets. The main idea is to densely sample feature points in each frame,
and track them in the video based on optical flow. Multiple descriptors are
computed along the trajectories of feature points to capture shape, appearance
and motion information. Interestingly, motion boundary histograms (MBH) \citep{Dalal2006}
give the best results due to their robustness to camera motion.

MBH is based on derivatives of optical flow, which is a simple and efficient
way to achieve robustness to camera motion. However, MBH only suppresses certain camera motions and, thus,
we can benefit from explicit camera motion estimation. Camera motion
generates many irrelevant trajectories in the background in realistic
videos. We can prune them
and only keep trajectories from humans and objects of interest, if we know the
camera motion, see Figure~\ref{fig:compare_flow}. Furthermore, given the camera
motion, we can correct the optical flow, so that the motion vectors from human body
are independent of camera motion. This improves the performance of motion descriptors
based on optical flow, \ie, HOF (histograms of optical flow) and MBH. We illustrate
the difference between the original and corrected optical flow in the middle two
columns of Figure \ref{fig:compare_flow}.

Besides improving low-level video descriptors, we also employ Fisher
vectors \citep{Sanchez2013} to encode local descriptors into a holistic
representation. Fisher vectors have been shown to give
superior performance over bag-of-words in image classification \citep{Chatfield2011,Sanchez2013}.
Our experimental results prove that the same conclusion also holds for
a variety of  recognition tasks in the video domain.

We consider three challenging problems to demonstrate the effectiveness of
our proposed framework. First, we consider the classification of basic action
categories using six of the most challenging datasets. Second, we consider
the localization of actions in feature length movies, including four action classes:
\emph{drinking, smoking, sit down}, and \emph{open door} from \citep{Duchenne2009,Laptev2007}.
Third, we consider classification of more high-level complex event categories
using the \trecvid MED 2011 dataset \citep{over2012}.

On all three tasks we obtain state-of-the-art performance, improving over earlier
work that relies on combining more feature channels, or using more complex models.
For action localization in full length movies, we also propose a modified
non-maximum-suppression technique that avoids a bias towards selecting short
segments, and further improves the detection performance.
This paper integrates and extends our previous results which have appeared
in earlier papers \citep{Dan2013,Wang2013a}.
The code to compute improved trajectories and descriptors is available
online.\footnote{\scriptsize \url{http://lear.inrialpes.fr/~wang/improved_trajectories}\label{fn:code}}

The rest of the paper is organized as follows. \sect{related} reviews related
work. We detail our improved trajectory features by explicit camera motion estimation in
\sect{improve}. Feature encoding and non-maximum-suppression for action
localization are presented in \sect{encoding} and \sect{nms}.
Datasets and evaluation protocols are described in \sect{datasets}.
Experimental results are given in \sect{experiments}. Finally, we present our conclusions in \sect{conclusion}.

\section{Related work}
\label{sec:related}

Feature trajectories \citep{Matikainen2009,Messing2009,Sun2009,Wang2013} have
been shown to be a good way for capturing the intrinsic dynamics of video
data. Very few approaches consider camera motion when extracting feature
trajectories for action recognition. \citet{Uemura2008} combine feature matching with image segmentation to estimate the dominant camera motion, and then separate feature tracks from the background. \citet{Wu2011b} apply a low-rank assumption to decompose feature trajectories into camera-induced and object-induced components. \citet{Gaidon2013} use efficient image-stitching techniques to compute the approximate motion of the background plane and generate stabilized videos before extracting dense trajectories \citep{Wang2013} for activity recognition.

Camera motion has also been considered in other types of video representations. \citet{ikizler10eccv} use of a homography-based
motion compensation approach in order to estimate the foreground optical flow field. \citet{li2012} recognize different camera motion types such as pan, zoom and tilt to separate foreground and background motion for video retrieval and summarization. Recently, \citet{Park2013} perform weak stabilization to remove both camera and object-centric motion using coarse-scale optical flow for pedestrian detection and pose estimation in video.

Due to the excellent performance of dense trajectories on a wide range of action datasets \citep{Wang2013}, there are several approaches try to improve them from different perspectives. \citet{Vig2012} propose to use saliency-mapping algorithms to prune background features. This results in a more compact video representation, and improves action recognition accuracy. \citet{Jiang2012} cluster dense trajectories, and use the cluster centers
as reference points so that the relationship between them can be modeled.
\citet{Jain2013} decompose visual motion into dominant and residual motions both for extracting trajectories and computing descriptors.

Besides carefully engineering video features, some recent work explores learning low-level features from video data \citep{Le2011,yang12eccv}. 
For example, \citet{cao12eccv} consider feature pooling based on scene-types, where video frames are assigned to scene types and their features are aggregated in the corresponding scene-specific representation. Along similar lines, \citet{ikizler10eccv} combines local person and object-centric features, as well as global scene  features. Others not only include object detector responses, but also use speech recognition, and character recognition systems to extract additional high-level features \citep{Natarajan2012}.

A complementary line of work has focused on considering more sophisticated models for action recognition that go beyond simple bag-of-words representations,
and aimed to explicitly capture the spatial and temporal structure of actions, see \eg, \citep{Gaidon2011,matikainen10eccv}. Other authors have focused on explicitly modeling interactions between people and objects, see \eg, \citep{gupta09pami,prest13pami}, or used multiple instance learning to suppress irrelevant background features \citep{sapienza2012}. Yet others have used graphical model structures to explicitly model the presence of sub-events \citep{izadinia12eccv,Tang2012}. \citet{Tang2012} use a variable-length discriminative HMM model which infers latent sub-actions together with a non-parametric duration distribution. \citet{izadinia12eccv} use a tree-structured CRF to model co-occurrence relations among sub-events and complex event categories, but require additional labeling of the sub-events unlike \citet{Tang2012}.

Structured models for action recognition seem promising to model basic actions such as \emph{drinking, answer phone,} or \emph{get out of car},
which could be decomposed into more basic action units, \eg, the ``actom'' model of \citet{Gaidon2011}. However, as the definition of the category becomes more high-level, such as \emph{repairing a vehicle tire}, or \emph{making a sandwich}, it becomes less clear to what degree it is possible to learn the structured models from limited amounts of training data, given the much larger amount of intra-class variability.
Moreover, more complex structured models are generally more computationally demanding, which limits their usefulness in large-scale settings.
To sidestep these potential disadvantages of more complex models, we instead explore the potential of recent advances in robust feature pooling strategies developed in the object recognition literature.

In particular, in this paper we explore the potential of the Fisher vector encoding \citep{Sanchez2013} as a robust feature pooling technique that has been proven to be among the most effective for object recognition \citep{Chatfield2011}.
While recently FVs have been explored by others for action recognition \citep{Sun2013,wang12accv},
we are the first to use them in a large, diverse, and comprehensive evaluation.
In parallel to this paper, \citet{Jain2013} complemented the dense trajectory descriptors with new features computed from optical flow, and encoded them using vectors of 
locally aggregated descriptors (VLAD; \citealp{Jegou2011}), a simplified version of the Fisher vector. We compare to these works in our experimental evaluation.

\section{Improving dense trajectories} \label{sec:improve}

In this section, we first briefly review the dense trajectory features~\citep{Wang2013}. We, then, detail the major steps of our improved trajectory features including camera motion estimation, removing inconsistent matches using human detection, and extracting improved trajectory features, respectively.

\subsection{Dense trajectory features} \label{sec:extract_dtf}

The dense trajectory features approach~\citep{Wang2013} densely samples feature points for several spatial scales. Points in homogeneous areas are suppressed, as it is impossible to track them reliably. Tracking points is achieved by median
filtering in a dense optical flow field~\citep{Farneback2003}. In order
to avoid drifting, we only track the feature points for 15 frames and sample new points to replace them. We remove static feature trajectories as they do not contain motion information, and also prune trajectories with sudden large displacements.

\begin{figure}[t]
\centering
\includegraphics[width=0.95\linewidth]{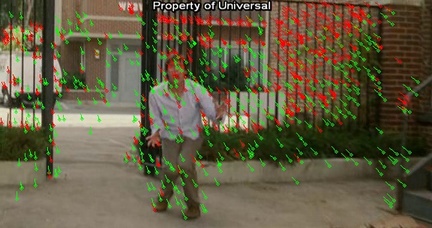}
\caption{Visualization of inlier matches of the estimated homography. Green arrows correspond to SURF descriptor matches, and red ones are from dense optical flow.}
\label{fig:surf_flow} 
\end{figure}

\begin{figure*}[!htbp]
\centering
\newcommand{\img}[1]{\includegraphics[width=0.245\linewidth,height=0.14\linewidth]{figures_rm_track_#1.jpg}}
\begin{tabular}{c@{\hspace{0.1em}}c@{\hspace{0.1em}}c@{\hspace{0.1em}}c}
\img{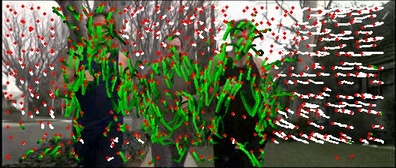} & \img{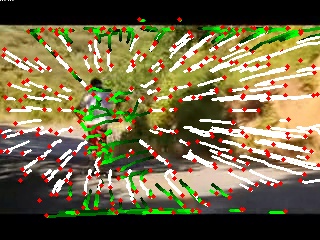} & \img{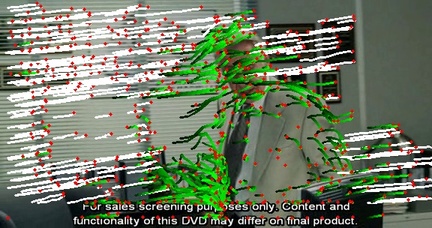} & \img{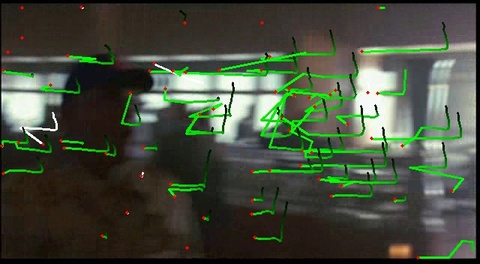} \vspace{-0.25em} \\
\img{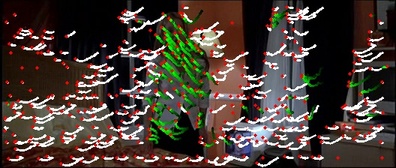} & \img{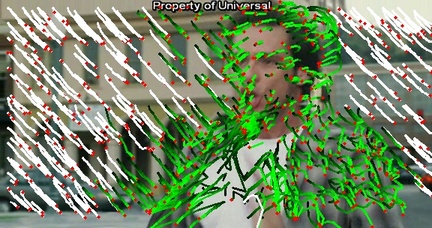} & \img{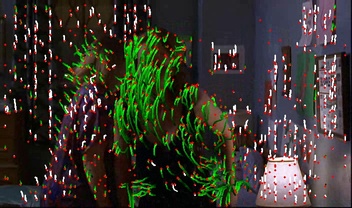} & \img{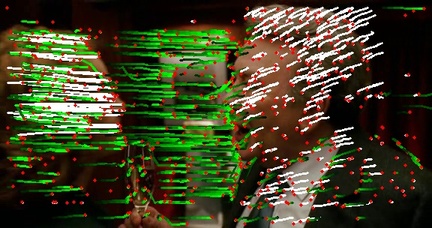} \vspace{-0.25em} \\
\end{tabular}
\vspace{0.5em}
\caption{Examples of removed trajectories under various camera motions, \eg, pan,
zoom, tilt. White trajectories are considered due to camera motion.
The red dots are the feature point positions in the current frame.
The last column shows two failure cases. The top one is due to severe motion blur.
The bottom one fits the homography to the moving humans as they dominate the whole frame.
} \label{fig:rm_track}
\end{figure*}

For each trajectory, we compute HOG, HOF and MBH descriptors with exactly the same parameters as in~\citep{Wang2013}. Note that we do not use the trajectory descriptor as it does not improve the overall performance significantly. All three descriptors are computed in the space-time volume aligned with the trajectory. HOG~\citep{Dalal2005} is based on the orientation of image gradients and captures the static appearance information. Both HOF~\citep{Laptev2008} and MBH~\citep{Dalal2006} measure motion information, and are based on optical flow. HOF directly quantizes
the orientation of flow vectors. MBH splits the optical flow into horizontal and vertical components, and quantizes the derivatives of each component.
The final dimensions of the descriptors are 96 for HOG, 108 for HOF and 2 $\times$ 96 for the two  MBH channels.

To normalize the histogram-based descriptors, \ie, HOG, HOF and MBH, we apply the recent RootSIFT~\citep{Arandjelovic2012} approach, \ie, square root each dimension after $\ell_1$ normalization. We do not perform $\ell_2$ normalization as in~\citep{Wang2013}.
This slightly improves the results without introducing additional computational cost.

\subsection{Camera motion estimation} \label{sec:camera_motion}

To estimate the global background motion, we assume that two consecutive frames are related by a homography~\citep{Szeliski2006}. This assumption holds in most cases as the global motion between two frames is usually small. It excludes
independently moving objects, such as humans and vehicles.

To estimate the homography, the first step is to find the correspondences between two frames. We combine two approaches in order to generate sufficient and complementary candidate matches. We extract speeded-up robust features (SURF; \citealp{Bay2006}) and match them based on the nearest neighbor rule. SURF features are obtained by first detecting interest points based on an approximation of the Hessian matrix and then describing them by a distribution of Haar-wavelet responses. The reason for choosing SURF features is their robustness to motion blur, as shown in a recent evaluation~\citep{Gauglitz2011}.

We also sample motion vectors from the optical flow, which provides us with dense matches between frames. Here, we use an efficient optical flow algorithm based on polynomial expansion~\citep{Farneback2003}. We select motion vectors for salient feature points using the good-features-to-track criterion~\citep{Shi1994}, \ie, thresholding the smallest eigenvalue of the autocorrelation matrix. Salient feature points are usually 
reproducible (stable under local and global perturbations, such as illumination variations or geometric transformation) and
distinctive (with rich local structure information).
Motion estimation on salient points is more reliable.

\begin{figure*}[t] 
\centering
\newcommand{\img}[1]{\includegraphics[width=0.245\linewidth,height=0.14\linewidth]{figures_human_detect_#1.jpg}}
\begin{tabular}{c@{\hspace{0.1em}}c@{\hspace{0.1em}}c@{\hspace{0.1em}}c}
\img{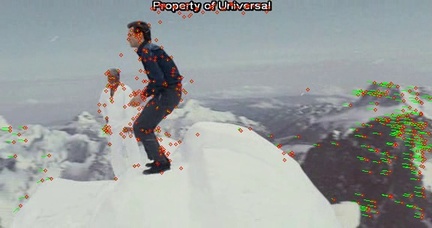} & \img{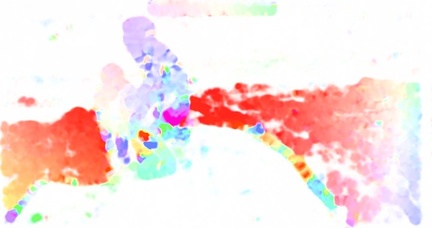} & \img{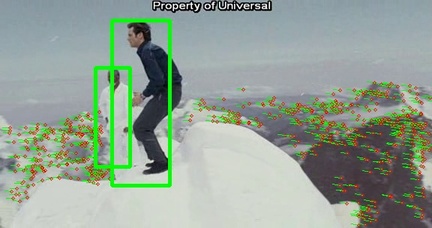} & \img{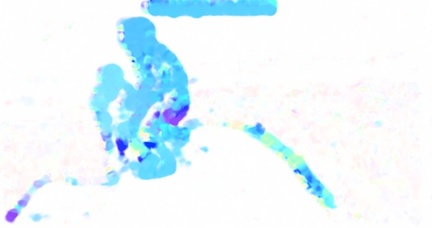} \vspace{-0.25em} \\
\img{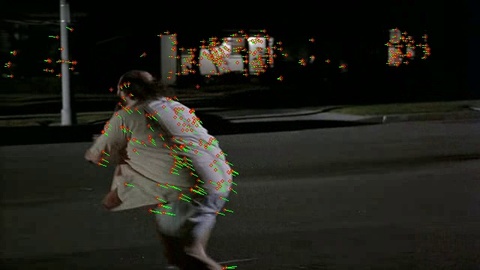} & \img{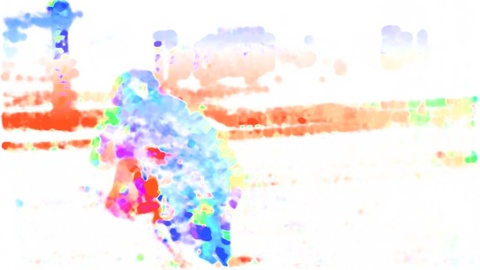} & \img{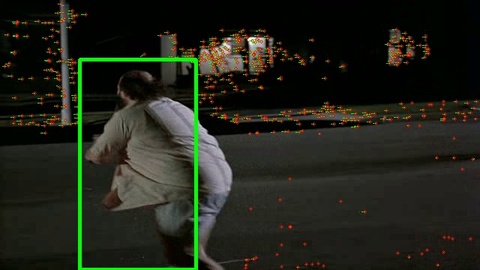} & \img{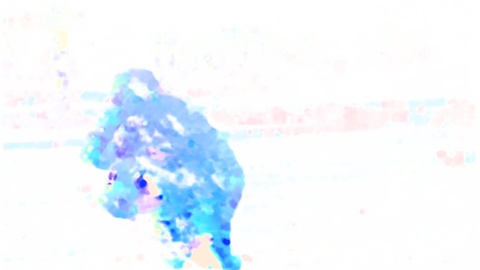} \vspace{-0.25em} \\
\img{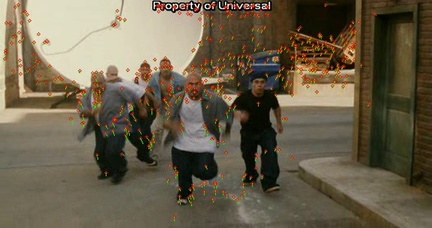} & \img{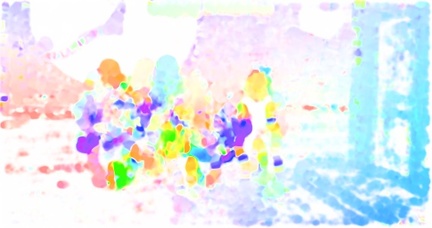} & \img{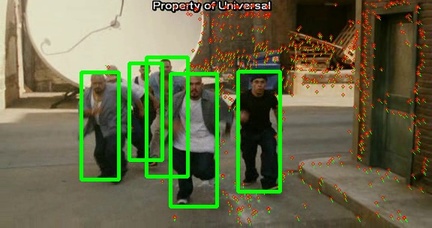} & \img{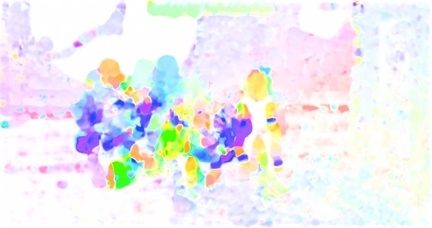} \vspace{-0.25em} \\
\end{tabular}
\vspace{0.5em}
\caption{Homography estimation without human detector (left) and with human detector (right). We show inlier matches in the first and third columns. The optical flow (second and fourth columns) is warped with the corresponding homography. The first and second rows show a clear improvement of the estimated homography when using a human detector. The last row presents a failure case. See the text for details.}
\label{fig:human_detect} 
\end{figure*}

The two approaches are complementary. SURF focuses on blob-type structures,
whereas~\citep{Shi1994} fires on corners and edges. Figure \ref{fig:surf_flow} visualizes the two types of matches in different colors. Combining them results in a more balanced distribution of matched points, which is critical for a good homography estimation.

We, then, estimate the homography using the random sample consensus method (RANSAC; \citealp{Fischler1981}).
RANSAC is a robust, non-deterministic algorithm for estimating the parameters of a model.
At each iteration it randomly samples a subset of the data to estimate the parameters of the model and computes the number of inliers that fit the model.
The final estimated parameters are those with the greatest consensus.
We then rectify the image using the homography to remove the camera motion.
Figure~\ref{fig:compare_flow} (two columns in the middle) demonstrates the difference of optical flow before and after rectification.
Compared to the original flow (the second column), the rectified version (the third column) suppresses the background camera motion and enhances the foreground moving objects.

For trajectory features, there are two major advantages of canceling out camera motion from optical flow. First, the motion descriptors can directly benefit from this. As shown in~\citep{Wang2013}, the performance of the HOF descriptor degrades significantly in the presence of camera motion. Our experimental results in \sect{experiments-action-recognition} show that HOF can achieve similar performance as MBH when we have corrected the optical flow. The combination of HOF and MBH can further improve the results as they represent zero-order (HOF) and first-order (MBH) motion information.

Second, we can remove trajectories generated by camera motion. This can be achieved by thresholding the displacement vectors of the trajectories in the warped flow field. If the displacement is very small, the trajectory is considered to be too similar to camera motion, and thus removed. Figure~\ref{fig:rm_track} shows examples of removed background trajectories. Our method works well under various camera motions (such as  pan, tilt and zoom) and only trajectories related to human actions are kept (shown in green in Figure \ref{fig:rm_track}). This gives us similar effects as sampling features based on visual saliency maps~\citep{Mathe2012,Vig2012}.

The last column of Figure \ref{fig:rm_track} shows two failure cases. The top one is
due to severe motion blur, which makes both SURF descriptor matching and optical
flow estimation unreliable. Improving motion estimation in the presence of motion
blur is worth further attention, since blur often occurs in realistic datasets.
In the bottom example, humans dominate the frame, which causes homography estimation to fail. We discuss a solution for the latter case below.

\subsection{Removing inconsistent matches due to humans} \label{sec:reject_outlier}

In action datasets, videos often focus on the humans performing the action. As a
result, it is very common that humans dominate the frame, which can be a problem
for camera motion estimation as human motion is in general not consistent with it.
We propose to use a human detector to remove matches from human regions.
In general, human detection in action datasets is rather difficult, as humans appear in
many different poses when performing the action. Furthermore, the person could  be only partially visible  due to occlusion or being partially out of view. 

Here, we apply a state-of-the-art human detector~\citep{Prest2012a}, which adapts the general part-based human detector~\citep{Felzenszwalb2010} to action datasets. The detector combines several part detectors dedicated to different regions of the human body (including full person, upper-body and face). It is trained using
the PASCAL VOC07 training data for humans as well as near-frontal upper-bodies from~\citep{Ferrari2008}.
We set the detection  threshold to 0.1. If the confidence of a detected window is higher than that, we consider it to be a positive sample. This is  a high-recall operating point where few human detections are missed. 
Figure \ref{fig:human_detect}, third column, shows some examples of human detection results.

We use the human detector as a mask to remove feature matches inside the bounding
boxes when estimating the homography. Without human detection (the left two columns of Figure \ref{fig:human_detect}), many features from the moving humans become inlier matches and the homography is, thus, incorrect. As a result, the corresponding optical flow is not correctly warped. In contrast, camera motion is successfully compensated (the right two columns of Figure \ref{fig:human_detect}), when the human bounding boxes are used to remove matches not corresponding to camera motion. The last row of Figure \ref{fig:human_detect} shows a failure case. The homography does not fit the background very well despite detecting the humans correctly, as the background is represented by two planes, one of which is very close to the camera. In our experiments we compare the performance  with and without human detection.

\begin{figure}[t] 
\centering
\newcommand{\img}[1]{\includegraphics[width=0.48\linewidth,height=0.3\linewidth]{figures_human_detector_#1.jpg}}
\begin{tabular}{c@{\hspace{0.1em}}c}
\img{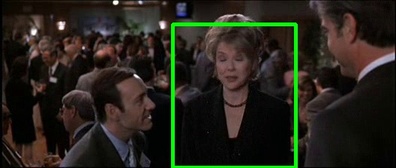} & \img{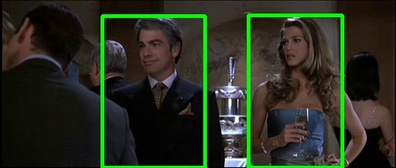} \vspace{-0.25em} \\
\img{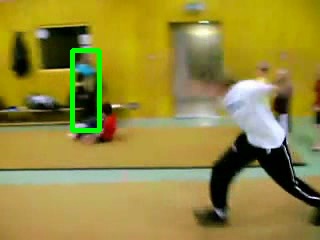} & \img{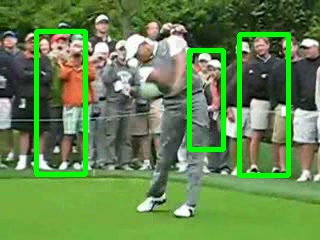} \vspace{-0.25em} \\
\img{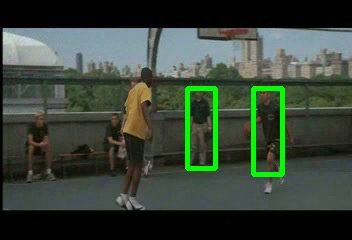} & \img{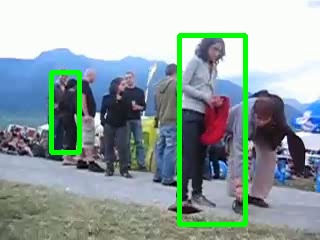} \vspace{-0.25em} \\
\end{tabular}
\vspace{0.5em}
\caption{Examples of human detection results. The first row is from Hollywood2, whereas the last two rows are from HMDB51.
Not all humans are detected correctly as human detection on action datasets is very challenging.} \label{fig:human_detector}
\end{figure}

The human detector does not always work perfectly. In Figure \ref{fig:human_detector}, we show some failure cases, which are typically due to complex human body poses, self occlusion, motion blur \etc. 
In order to compensate for missing detections, we track all the bounding boxes obtained by the human detector. Tracking is performed forward and backward for each frame of the video. Our approach is simple: we take the average
motion vector~\citep{Farneback2003} and propagate the detections to the
next frame. We track each bounding box for at most 15 frames and stop
if there is a 50\% overlap with another bounding box.   All the human
bounding boxes are available online.\footnote{\scriptsize \url{http://lear.inrialpes.fr/~wang/improved_trajectories}\label{fn:code}} In the
following, we always use the human detector to remove potentially
inconsistent matches before computing the homography, unless stated
otherwise.

\subsection{Improved trajectory features} \label{sec:extract_itf}

To extract our improved trajectories, we sample and track feature points exactly the same way as in~\citep{Wang2013}, see \sect{extract_dtf}.
To compute the descriptors, we first estimate the homography with RANSAC using the feature matches extracted between each pair of consecutive frames; matches on detected humans are removed.
We warp the second frame with the estimated homography.
Homography estimation takes around 5 milliseconds for each pair of frames.
The optical flow~\citep{Farneback2003} is then
re-computed between the  first and the warped second frame. Motion descriptors (HOF and MBH) are computed on the warped optical flow. The HOG descriptor remains unchanged. We estimate the homography and warped optical flow for every two frames independently to avoid error propagation. We use the same parameters and the RootSIFT normalization as the baseline described in section~\ref{sec:extract_dtf}.
We further utilize these stabilized motion vectors to remove background trajectories. For each trajectory, we compute the maximal magnitude of the motion vectors during its length of 15 frames. If the maximal magnitude is lower than a threshold (set to one pixel, \ie, the motion displacement is less than one pixel between each pair of frames), the trajectory is considered to be consistent with camera motion, and thus removed.


%

\section{Feature encoding}
\label{sec:encoding}

In this section, we present how we aggregate local descriptors into a holistic representation, and augment this representation with weak spatio-temporal location information.

\subsection{Fisher vector}


The Fisher vector (FV; \citealp{Sanchez2013}) was found to be the most effective encoding technique in a recent evaluation study of feature pooling techniques for object recognition \citep{Chatfield2011};
this evaluation included also bag-of-words (BOW), sparse coding techniques, and several variants.
The FV extends the BOW representation as it encodes both first and second order statistics between the video descriptors and a diagonal covariance Gaussian mixture model (GMM).
Given a video, let $x_n\in \R^D$ denote the $n$-th $D$-dimensional video descriptor,
$q_{nk}$ the soft assignment of $x_n$ to the $k$-th Gaussian,
and $\pi_k$, $\mu_k$ and $\sigma_k$ are the weight, mean, and diagonal of the covariance matrix of the $k$-th Gaussian respectively.
After normalization with the inverse Fisher information matrix (which renders the  FV invariant to the parametrization), the $D$-dimensional gradients \wrt the mean and variance of the $k$-th Gaussian are given by:
\bea
G_{\mu_k} & = & \sum_{n=1}^N q_{nk} \left[x_n-\mu_k\right] / \sqrt{\sigma_k\pi_k},\label{eq:FVm}\\
G_{\sigma_k} & = & \sum_{n=1}^N q_{nk} \left[(x_n-\mu_k)^2 -\sigma^2_k\right] / \sqrt{2\sigma_k^2\pi_k}.\label{eq:FVs}
\eea

For each descriptor type $x_n$, we can represent the video as a $2DK$ dimensional Fisher vector.
To compute FV, we first reduce the descriptor dimensionality by a factor of two using principal component analysis (PCA), as in~\citep{Sanchez2013}.
We then randomly sample a subset of $1000\times K$ descriptors from the training set to estimate a GMM with $K$ Gaussians.
After encoding the descriptors using Eq. \eqref{eq:FVm} and \eqref{eq:FVs}, we apply power and $\ell_2$ normalization to the final Fisher vector representation as in~\citep{Sanchez2013}.
A linear SVM is used for classification.


Besides FV, we also consider BOW histograms as a baseline for feature encoding. We use the soft assignments to the same Gaussians as used for the FV instead of hard assignment with k-means clustering \citep{gemert10pami}. Soft assignments have been reported to yield better performance, and since the same GMM vocabulary is used as for the FV, it also rules out any differences due to the vocabulary. 
For BOW, we consider both linear and RBF-$\chi^2$ kernel for the SVM classifier. In the case of linear kernel, we employ the same power and $\ell_2$ normalization as FV, whereas $\ell_1$ normalization is used for RBF-$\chi^2$ kernel.

To combine different descriptor types, we encode each descriptor type separately and concatenate their normalized BOW or FV representations together. In the case of multi-class classification, we use a one-against-rest approach and select the class with the highest score. For the SVM hyperparameters, we set the class weight $w$ to be inversely proportional to the number of samples in each class so that both positive and negative classes contribute equally in the loss function. We set the regularization parameter $C$ by cross validation on the training set, by testing values in the range   $C\in\{3^{-2}, 3^{-1},\cdots, 3^7\}$. In all experiments, we use the same settings.




\subsection{Weak spatio-temporal location information}
\label{sec:spm}

To go beyond a completely orderless representation of the video content in a BOW histogram or FV, we consider including a weak notion of spatio-temporal location information of the local features. For this purpose, we use the spatio-temporal pyramid (STP) representation~\citep{Laptev2008}, and compute separate BOW or FV over cells in spatio-temporal grids. We also consider the spatial Fisher vector (SFV) of \citep{krapac11iccv}, which computes per visual word the mean and variance of the 3D spatio-temporal location of the assigned features. This is similar to extending the feature vectors (HOG, HOF or MBH) with the 3D locations, as done in~\citep{mccann12accv,sanchez12prl}; the main difference being that the latter do clustering on the extended feature vectors while this is not the case for the SFV.
SFV is also computed in each cell of STP. To combine SFV with BOW or FV, we simply concatenate them together. 



\section{Non-maximum-suppression for localization}
\label{sec:nms}

For the action localization task we employ a temporal sliding window approach.
We score a large pool of candidate detections that are obtained by sliding windows of various lengths across the video.
Non-maximum suppression (NMS) is performed to delete windows that have an overlap greater than 20\% with higher scoring windows.
In practice, we use candidate windows of length $30,60,90$, and $120$ frames, and slide the windows in steps of $30$ frames.

\begin{figure}
\begin{center}
\includegraphics[width=0.6\linewidth]{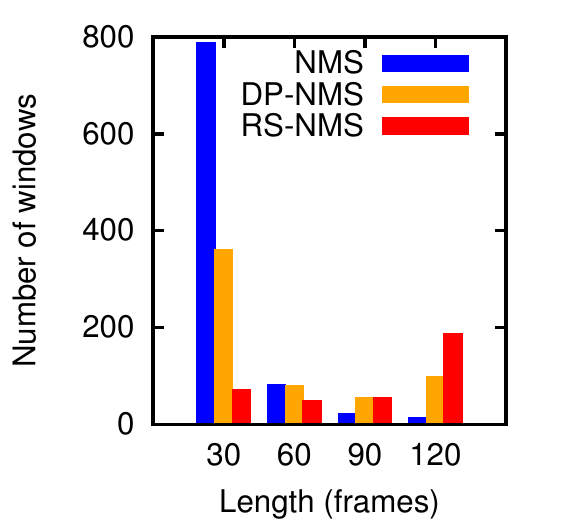}
\end{center}
\caption{%
Histograms of the window sizes on the \emph{Coffee and Cigarettes} dataset after three variants of non-maxima suppression: classic non-maximum suppression (NMS), dynamic programming non-maximum suppression (DP-NMS), and re-scored non-maximum suppression (RS-NMS). 
Two of the methods, NMS and DP-NMS, select mostly short windows, 30-frames long, while the RS-NMS variant sets a bias towards longer windows, 120-frames long. In practice we prefer longer windows as they tend to cover better the action.
}
\label{fig:nms-variants-histograms}
\end{figure}

Preliminary experiments showed that there is a strong tendency for the  NMS  to retain short windows, see \fig{nms-variants-histograms}.
This is due to the fact that if a relatively long action appears, it is likely that there are short sub-sequences that just contain the most characteristic features for the action. Longer windows might better cover the action, but are likely to include less characteristic features as well (even if they lead to positive classification by themselves),  and might include background features due to imperfect temporal alignment.

To address this issue we consider re-scoring the segments by multiplying their score with their duration, before applying NMS (referred to as RS-NMS). We also consider a variant where the goal is to select a subset of candidate windows that (i)~covers the entire video, (ii)~does not have overlapping windows, and (iii)~maximizes the sum of scores of the selected windows. We formally express this method as an optimization problem:
\begin{eqnarray}
         \underset{y}{\maximize} &  & \sum_{i=1}^n y_i s_i \label{eq:dp}\\
         \text{subject to}       &  & \bigcup_{i:y_i=1} l_i = T, \nonumber \\
                                 &  & \forall_{y_i=y_j=1}: l_i \cap l_j = \emptyset, \nonumber \\
                                 &  & y_i \in \{0, 1\}, \; i = 1,\ldots, n.\nonumber 
\end{eqnarray}
where the boolean variables $y_1,\ldots,y_n$ represent the subset; $s_i$ and $l_i$ denote the score and the interval of window~$i$; $n$ is the total number of windows; $T$ is the interval that spans the whole video.

\begin{figure}
\begin{center}
\includegraphics[width=0.8\linewidth]{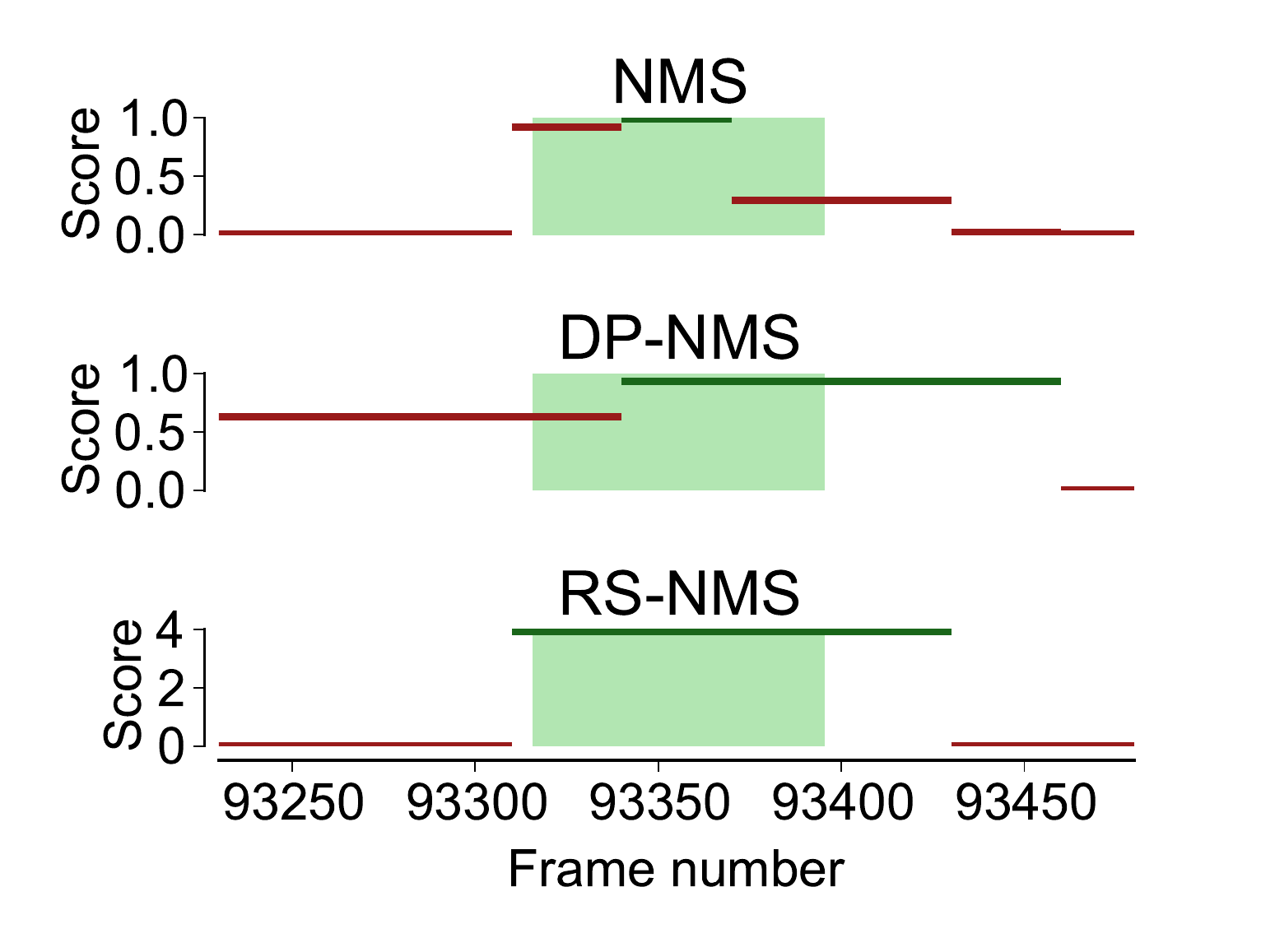}
\end{center}
\caption{%
Windows retained by NMS variants, green if they overlap more than $20\%$ with the true positive, red otherwise.
The green region denotes the ground-truth action.
For the NMS, the segments selected are too short.
The DP-NMS selects longer segments, but it does not align well with the true action as it maximizes the total score over the whole video.
The RS-NMS strikes a good balance of the segment's length and their score, and it gives the best solution in this example.}
\label{fig:nms-variants-example}
\end{figure}


The optimal subset is found efficiently by dynamic programming as follows.
We first divide the temporal domain into discrete time steps.
With each time step we associate a latent state: the temporal window that contains that particular time step. Each window is characterized by its starting point and duration.
A pairwise potential is used to enforce the first two constraints (full duration coverage and non-overlapping segments):
if a segment is not terminated at the current time step, the next time step should still be covered by the current segment, otherwise a new segment should be started.
We maximize the score based on an unary potential that is defined as the score of the associated time step.
The dynamic programming Viterbi algorithm is used to compute the optimal solution for the optimization problem of Equation (\ref{eq:dp}) using a forwards and backwards pass over the time steps. The runtime is linear in the number of time steps. 
We refer to this method as DP-NMS.

\fig{nms-variants-histograms} shows the histogram of durations of the windows that pass the non-maximum suppression stage using the different techniques, for the action \emph{smoking} used in our experiments in \sect{experiments-action-localization}.
The durations for the two proposed methods, DP-NMS and RS-NMS, have a more uniform distribution than that for the standard NMS method, with RS-NMS favouring the longest windows.
This behaviour is also observed in \fig{nms-variants-example}, which gives an example of the different windows retained for a specific video segment of the \emph{Coffee \& Cigarettes} movie.
DP-NMS selects longer windows than NMS, but they do not align well with the action and the score of the segments outside the action are high.
For this example, RS-NMS gives the best selection among the three methods, as it retains few segments and covers the action accurately.

%



\begin{figure*}[t!]
\centering \scriptsize
\newcommand{\img}[1]{\includegraphics[width=0.16\linewidth,height=0.12\linewidth]{figures_#1.jpg}}
\newcommand{\gap}{\hspace{0.2em}}
\begin{tabular}{c@{\gap}c@{\gap}c@{\gap}c@{\gap}c@{\gap}c}
    \img{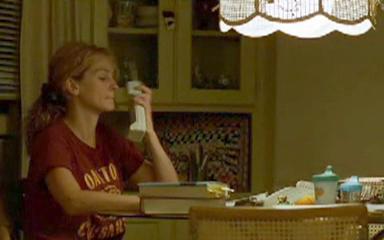} &
    \img{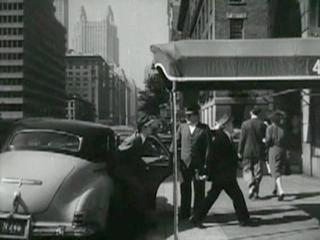} &
    \img{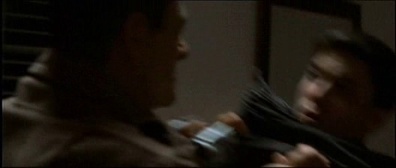} &
    \img{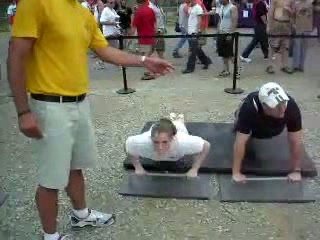} &
    \img{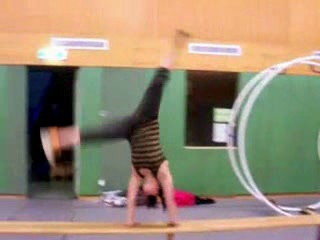} &
    \img{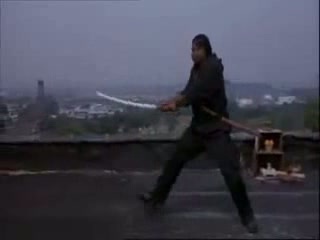} \\
    (a) answer-phone & (a) get-out-car & (a) fight-person & (b) push-up & (b) cartwheel & (b) sword-exercise \\

    \img{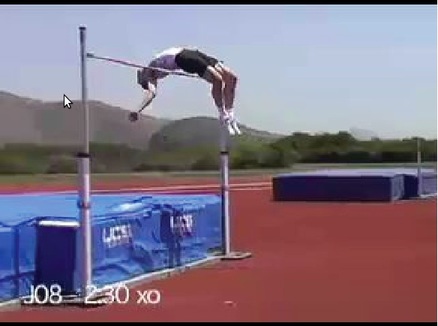} &
    \img{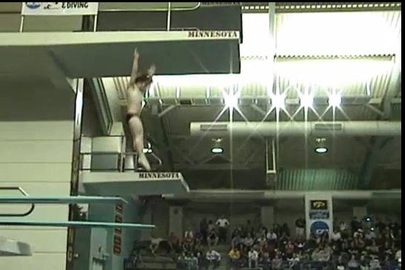} &
    \img{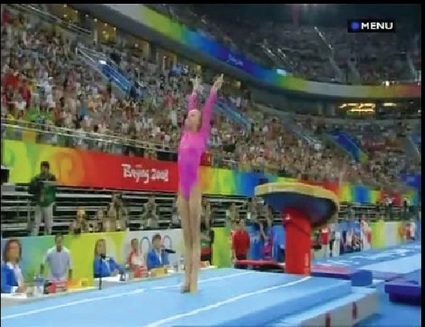} &
    \img{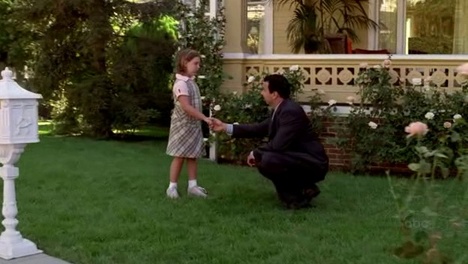} &
    \img{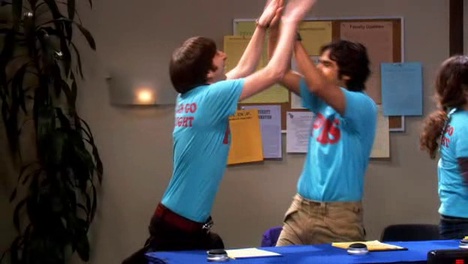} &
    \img{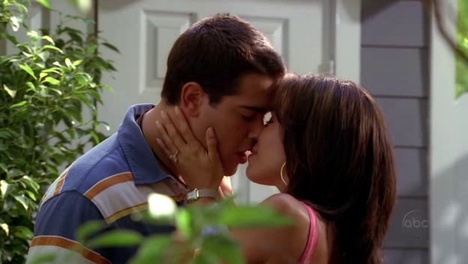} \\
    (c) high-jump & (c) spring-board & (c) vault & (d) hand-shake & (d) high-five & (d) kiss \\

    \img{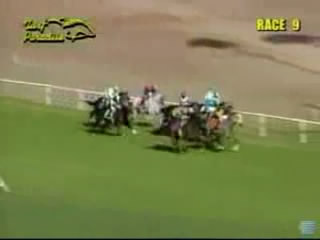} &
    \img{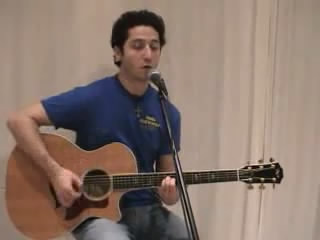} &
    \img{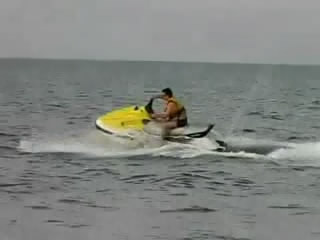} &
    \img{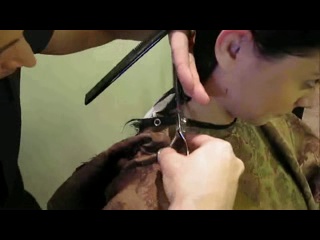} &
    \img{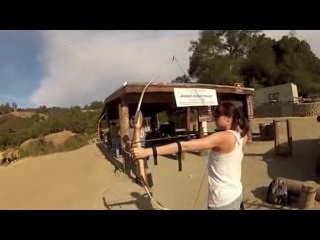} &
    \img{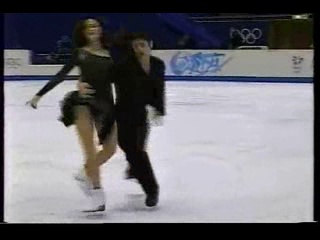} \\
    (e) horse-race & (e) playing-guitar & (e) ski-jet & (f) haircut & (f) archery & (f) ice-dancing \\
\end{tabular}

\renewcommand{\img}[1]{\includegraphics[width=0.24\linewidth,height=0.12\linewidth]{figures_localize_#1.jpg}}
\begin{tabular}{c@{\gap}c@{\gap}c@{\gap}c}
    \img{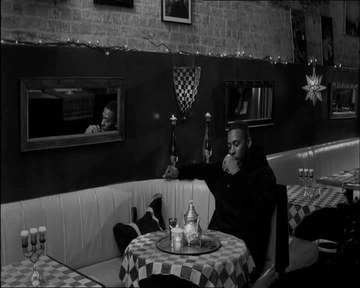} &
    \img{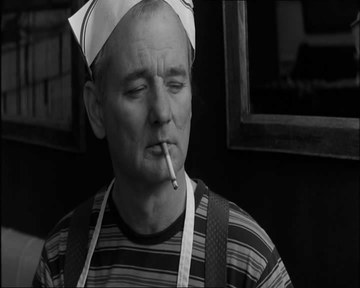} &
    \img{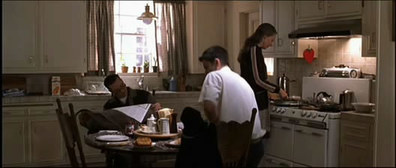} &
    \img{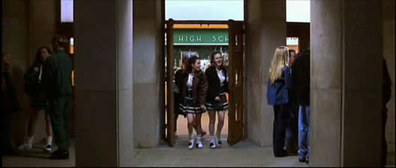} \\
    (g) drinking & (g) smoking & (h) sit-down & (h) open-door \\
\end{tabular}

\renewcommand{\img}[1]{\includegraphics[width=0.16\linewidth,height=0.12\linewidth]{figures_trecvid_#1.jpg}}
\begin{tabular}{c@{\gap}c@{\gap}c@{\gap}c@{\gap}c@{\gap}c}
    \img{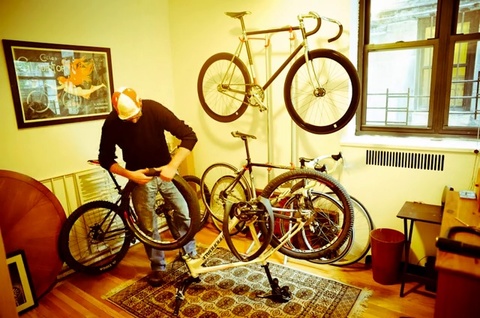} &
    \img{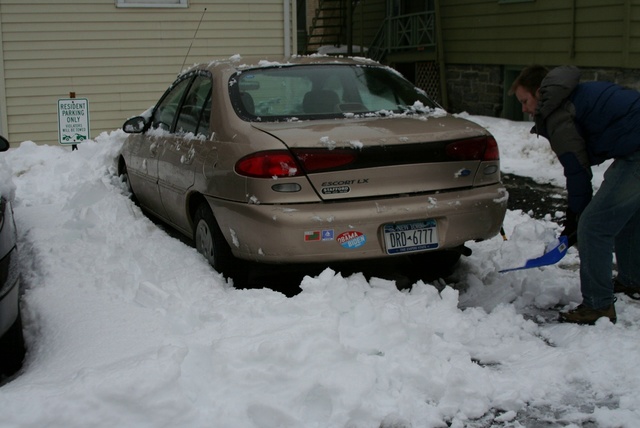} &
    \img{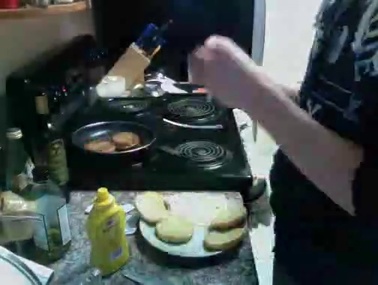} &
    \img{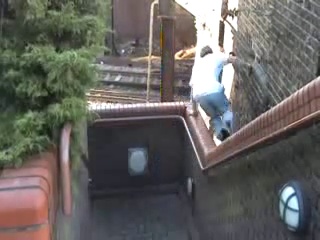} &
    \img{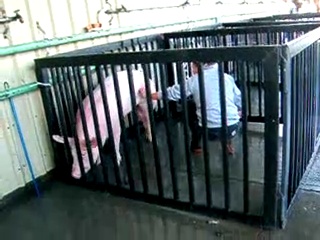} &
    \img{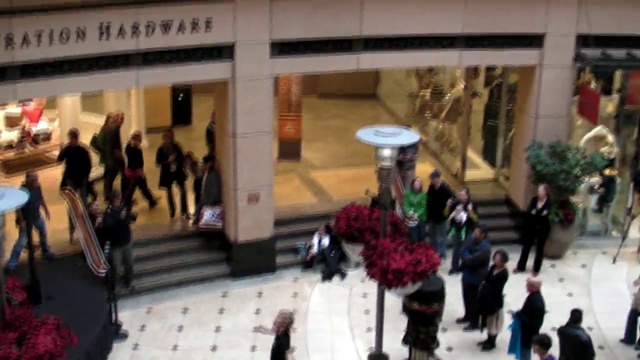} \\
    (i) changing-vehicle-tire & (i) unstuck-vehicle & (i) making-a-sandwich & (i) parkour & (i) grooming-an-animal & (i) flash-mob-gathering \\
\end{tabular}

\vspace{0.5em}
\caption{From top to bottom, example frames from (a) Hollywood2, (b) HMDB51, (c) Olympic Sports, (d) High Five, (e) UCF50, (f) UCF101, (g) \emph{Coffee and Cigarettes}, (h) DLSBP and (i) \trecvid MED 2011 . }
\label{fig:dataset}
\end{figure*}

\section{Datasets used for experimental evaluation}\label{sec:datasets}

In this section, we briefly describe the datasets and their evaluation protocols for the three tasks. We use six challenging datasets for action recognition (\ie, Hollywood2, HMDB51, Olympic Sports, High Five, UCF50 and UCF101), \emph{Coffee and Cigarettes} and DLSBP for action detection, and \trecvid MED 2011 for large scale event detection. In Figure~\ref{fig:dataset}, we show some sample frames from the datasets.

\subsection{Action recognition}
\label{subsec:action-recognition}

The {\bfseries Hollywood2} dataset~\citep{Marszalek2009} has been
collected from 69 different Hollywood movies and includes 12 action
classes.
It contains 1,707 videos split
into a training set (823 videos) and a test set (884 videos).
Training and test videos come from different
movies. The performance is measured by mean average precision (mAP)
over all classes, as in~\citep{Marszalek2009}.


The {\bfseries HMDB51} dataset~\citep{Kuehne2011} is collected from a
variety of sources ranging from digitized movies to YouTube
videos. 
In total, there are 51 action categories and 6,766 video sequences. We follow the original protocol
using three train-test splits~\citep{Kuehne2011}. For every class and split, there are 70 videos
for training and 30 videos for testing. We report average accuracy over the three splits as
performance measure. Note that in all the experiments we use the original
videos, not the stabilized ones.

The {\bfseries Olympic Sports} dataset~\citep{Niebles2010} consists of
athletes practicing different sports, which are collected from YouTube
and annotated using Amazon Mechanical Turk. There are 16 sports actions (such as high-jump,
pole-vault, basketball lay-up, discus), represented by a total of 783 video sequences. We use 649
sequences for training and 134 sequences for testing as recommended by the authors. We report
mAP over all classes, as in~\citep{Niebles2010}.

The {\bfseries High Five} dataset~\citep{Patron2010} 
consists of 300 video clips  extracted from 23 different TV shows.
Each of the clips contains one of four interactions: hand shake, high five, hug and kiss
(50 videos for each class). Negative examples (clips that don't contain any of the interactions)
make up the remaining 100 videos. Though the dataset is relatively small, it is challenging due to
large intra-class variation, and all the action classes are very similar to each other (\ie, interactions between two persons).
We follow the original setting in~\citep{Patron2010}, and compute average precision (AP) using a pre-defined two-fold cross-validation.

The {\bfseries UCF50} dataset~\citep{Reddy2012} has 50 action categories, consisting of real-world videos
taken from YouTube. The actions range from general sports to daily life exercises. For all 50 categories,
the videos are split into 25 groups. For each group, there are at least four action clips. In total,
there are 6,618 video clips. The video clips in the same group may share some common features,
such as the same person, similar background or viewpoint. We apply the leave-one-group-out
cross-validation as recommended in~\citep{Reddy2012} and report average accuracy over all classes.

The {\bfseries UCF101} dataset~\citep{Soomro2012} 
is extended from UCF50 with additional 51 action categories. In total, there are 13,320 video clips.
We follow the evaluation guidline from the THUMOS'13 workshop~\citep{THUMOS13} using three train-test splits.
In each split, clips from seven of the 25 groups are used as test samples, and the rest for training.
We report average accuracy over the three splits as performance measure.

%

\subsection{Action localization}
\label{subsec:action-localization}

The first dataset for action localization is extracted from the movie {\bfseries {Coffee and Cigarettes}}, and contains annotations for the actions \emph{drinking} and \emph{smoking}~\citep{Laptev2007}. The training set contains 41 and 70 examples  for each class respectively.
Additional training examples (32 and eight respectively) come from the movie \emph{Sea of Love}, and another 33 lab-recorded \emph{drinking} examples are included.
The test sets consist of about 20 minutes from \emph{Coffee and Cigarettes} for \emph{drinking}, with 38 positive examples;
for \emph{smoking} a sequence of about 18 minutes is used that contains 42 positive examples.

The {\bfseries DLSBP} dataset of Duchenne \etal~\citep{Duchenne2009} contains annotations for the actions \emph{sit down}, and \emph{open door}.
The training data comes from 15 movies, and contains 51 \emph{sit down} examples, and 38 for \emph{open door}.
The test data contains three full movies (\emph{Living in Oblivion}, \emph{The Crying Game}, and \emph{The Graduate}), which in total last for about 250 minutes, and contain 86 \emph{sit down}, and 91 \emph{open door} samples.

To measure performance we compute the average precision (AP) score as in \citep{Duchenne2009,Gaidon2011,Klaser2010b,Laptev2007}; considering a detection as correct when it overlaps (as measured by intersection over union) by at least 20\% with a ground truth annotation.


\subsection{Event recognition}
\label{subsec:event-recognition}

The {\bfseries \trecvid MED 2011} dataset~\citep{over2012} is the largest dataset we consider.
It consists of consumer videos from 15 categories that are more complex than the basic actions considered in the other datasets, \eg, \emph{changing a vehicle tire}, or \emph{birthday party}.
For each category between 100 and 300 training videos are available.
In addition, 9,600 videos are available that do not contain any of the 15 categories;
this data is referred to as the \emph{null} class.
The test set consists of 32,000 videos, with a total length of over 1,000 hours, and includes 30,500 videos of the \emph{null} class.



We follow two experimental setups in order to compare our system to previous work.
The first setup is the one described above, which was also used in the \trecvid 2011 MED challenge.
The performance is evaluated using average precision (AP) measure.
The second setup is the one of Tang \etal~\citep{Tang2012}.
They split the data into three subsets:
EVENTS, which contains 2,048 videos from the 15 categories, but doesn't include the \emph{null} class;
DEV-T, which contains 602 videos from the first five categories and the 9,600 \emph{null} videos;
and DEV-O, which is the standard test set of 32,000 videos.\footnote{The number of videos in each subset varies slightly from the figures reported in \citep{Tang2012}. The reason is that there are multiple releases of the data. For our experiments, we used the labels from the \texttt{LDC2011E42} release.}
As in \citep{Tang2012}, we train on the EVENTS set and report the performance in AP on the DEV-T set for the first five categories and on the DEV-O set for the remaining ten actions.

The videos in the \trecvid dataset vary strongly in size: durations range from a few seconds to one hour, while the resolution ranges from low quality $128\times88$ to full HD $1920\times1080$.
We rescale the videos to a width of at most $480$ pixels, preserving the aspect ratio,
and temporally sub-sample them by discarding every second frame in order to make the dataset computationally more tractable.
These rescaling parameters were selected on a subset of the MED dataset;
we present an exhaustive evaluation of the impact of the video resolution in  \sect{experiments-event-recognition}.
Finally, we also randomly sample the generated features to reduce the computational cost for feature encoding. This is done only for videos longer than 2000 frames, \ie,
the sampling ratio is set to 2000 divided by the total number of frames.




\setcounter{table}{0}
\begin{table*}  [!htbp]
  \centering
  \scalebox{0.95} {
  \begin{tabular}{|rc||c|c|c|c|c||c|c|c|c|c|}
    \hline
     &
     & \multicolumn{5}{c||}{Hollywood2}
     & \multicolumn{5}{c| }{HMDB51}
    \\
     &
     & \multicolumn{3}{c| }{Bag-of-words}
     & \multicolumn{2}{c||}{Fisher vector}
     & \multicolumn{3}{c| }{Bag-of-words}
     & \multicolumn{2}{c| }{Fisher vector}
    \\
     &
     & \multicolumn{1}{c| }{$\chi^2$ kernel}
     & \multicolumn{2}{c| }{linear kernel}
     & \multicolumn{2}{c||}{linear kernel}
     & \multicolumn{1}{c| }{$\chi^2$ kernel}
     & \multicolumn{2}{c| }{linear kernel}
     & \multicolumn{2}{c| }{linear kernel}
    \\
     $K$ & STP            &  BOW       &  BOW       &  BOW+SFV   &  FV        &  FV+SFV    &  BOW       &  BOW       &  BOW+SFV   &  FV        &  FV+SFV \\
   \hline\hline
      64 & ---            &  44.4\%    &  39.8\%    &  40.3\%    &  55.0\%    &  56.5\%    &  30.5\%    &  28.3\%    &  28.0\%    &  45.8\%    &  47.9\%    \\
      64 & H3             &  48.0\%    &  44.9\%    &  45.0\%    &  57.9\%    &  59.2\%    &  35.8\%    &  30.1\%    &  33.1\%    &  48.0\%    &  49.4\%    \\
      64 & T2             &  48.3\%    &  43.4\%    &  46.8\%    &  57.1\%    &  58.5\%    &  34.9\%    &  30.9\%    &  32.5\%    &  48.3\%    &  49.5\%    \\
      64 & T2$+$H3        &  50.2\%    &  46.8\%    &  46.4\%    &  59.4\%    &  59.5\%    &  37.1\%    &  32.5\%    &  34.2\%    &  50.3\%    &  51.1\%    \\
   \hline
     128 & ---            &  45.8\%    &  42.1\%    &  43.5\%    &  57.1\%    &  58.5\%    &  33.8\%    &  31.9\%    &  32.2\%    &  48.2\%    &  50.3\%    \\
     128 & H3             &  51.3\%    &  46.2\%    &  48.1\%    &  58.8\%    &  60.0\%    &  38.0\%    &  32.3\%    &  37.5\%    &  49.9\%    &  51.1\%    \\
     128 & T2             &  50.5\%    &  45.5\%    &  49.4\%    &  58.8\%    &  59.9\%    &  38.2\%    &  32.9\%    &  36.2\%    &  50.2\%    &  51.1\%    \\
     128 & T2$+$H3        &  52.4\%    &  48.4\%    &  48.2\%    &  61.0\%    &  60.7\%    &  40.5\%    &  35.8\%    &  37.9\%    &  51.9\%    &  52.6\%    \\
   \hline
     256 & ---            &  49.4\%    &  44.9\%    &  45.9\%    &  57.9\%    &  59.6\%    &  36.6\%    &  33.1\%    &  35.0\%    &  50.0\%    &  51.9\%    \\
     256 & H3             &  52.9\%    &  46.0\%    &  50.6\%    &  59.0\%    &  61.0\%    &  40.6\%    &  36.2\%    &  40.4\%    &  51.4\%    &  52.3\%    \\
     256 & T2             &  52.0\%    &  47.0\%    &  51.3\%    &  59.3\%    &  60.3\%    &  41.3\%    &  35.7\%    &  39.7\%    &  51.5\%    &  52.0\%    \\
     256 & T2$+$H3        &  53.6\%    &  50.2\%    &  50.2\%    &  61.0\%    &  61.3\%    &  43.5\%    &  39.2\%    &  41.2\%    &  52.6\%    &  53.2\%    \\
   \hline
     512 & ---            &  50.2\%    &  46.8\%    &  49.0\%    &  58.9\%    &  60.5\%    &  40.3\%    &  35.6\%    &  37.9\%    &  51.3\%    &  53.2\%    \\
     512 & H3             &  53.1\%    &  49.5\%    &  51.2\%    &  59.5\%    &  61.5\%    &  43.4\%    &  38.4\%    &  41.5\%    &  51.4\%    &  52.3\%    \\
     512 & T2             &  53.9\%    &  49.4\%    &  52.8\%    &  60.2\%    &  61.0\%    &  42.6\%    &  39.1\%    &  42.2\%    &  52.2\%    &  53.3\%    \\
     512 & T2$+$H3        &  55.5\%    &  51.6\%    &  51.3\%    &{\bf61.7\%} &{\bf61.9\%} &  45.2\%    &  42.1\%    &  43.5\%    &  52.7\%    &  53.7\%    \\
   \hline
    1024 & ---            &  52.3\%    &  48.5\%    &  50.4\%    &  58.9\%    &  60.9\%    &  42.3\%    &  39.2\%    &  39.9\%    &  51.4\%    &  53.9\%    \\
    1024 & H3             &  55.6\%    &  50.6\%    &  52.6\%    &  59.4\%    &  61.2\%    &  45.4\%    &  40.8\%    &  44.2\%    &  51.7\%    &  52.8\%    \\
    1024 & T2             &  54.6\%    &  52.0\%    &{\bf54.5\%} &  59.7\%    &  60.7\%    &  46.0\%    &  41.8\%    &{\bf46.3\%} &  52.5\%    &  53.0\%    \\
    1024 & T2$+$H3        &{\bf56.6\%} &{\bf52.9\%} &  53.5\%    &  61.2\%    &  61.8\%    &{\bf47.5\%} &{\bf43.9\%} &  45.7\%    &{\bf53.3\%} &{\bf53.8\%} \\
   \hline
  \end{tabular}
  }
	\caption{Comparison of bag-of-words and Fisher vectors using the non-stabilized MBH descriptor under different parameter settings. We use $\ell_1$ normalization for the $\chi^2$ kernel, and power and $\ell_2$ normalization for the linear kernel.}
  \label{tab:bov-vs-fv}
\end{table*}

\begin{figure*}[!htbp]
\centering
\newcommand{\img}[1]{\includegraphics[width=0.5\linewidth]{#1.pdf}}
\begin{tabular}{c@{\hspace{-0.3em}}c}
\img{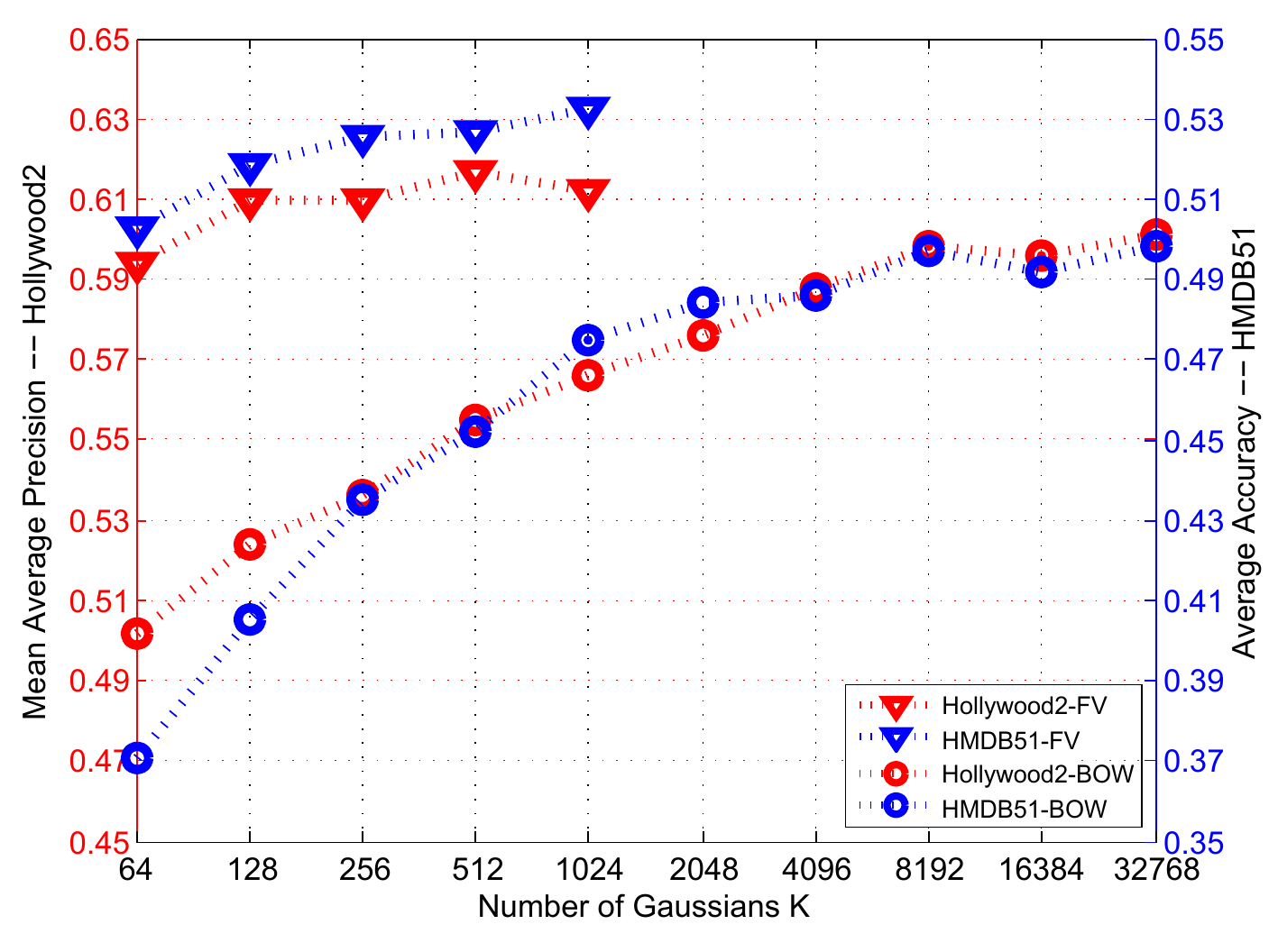} & \img{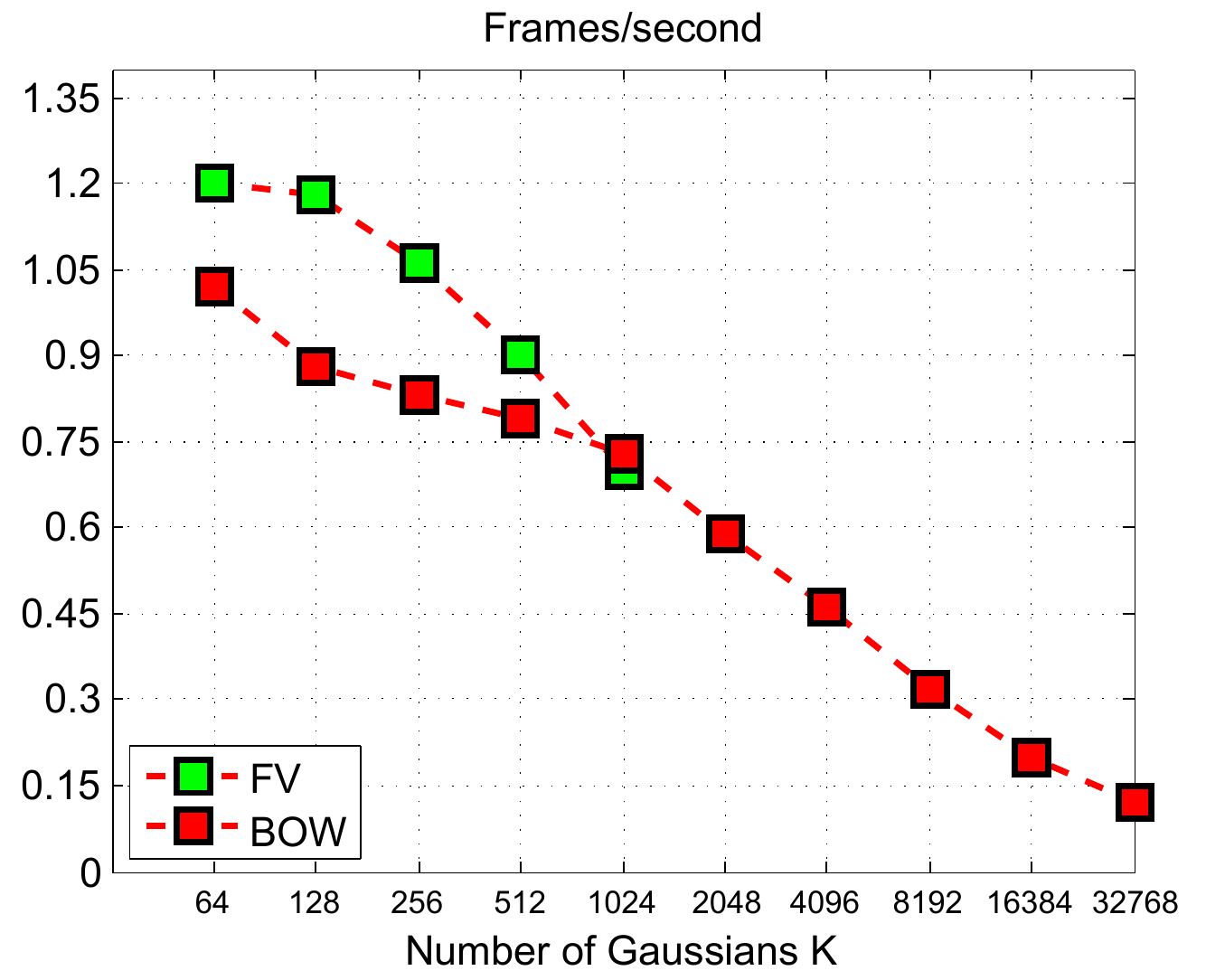} \vspace{-0.em}\\
\end{tabular}
\caption{Comparing BOW (RBF-$\chi^2$ kernel) using large vocabularies with FV (linear kernel). For both, we only use STP (T2$+$H3) without SFV. Left: performance on Hollywood2 and HMDB51. Right: runtime speed on a Hollywood2 video of resolution $720\times480$ pixels. }
\label{fig:large_k} 
\end{figure*}

\section{Experimental results}
\label{sec:experiments}

Below, we present our experimental evaluation results for action recognition in \sect{experiments-action-recognition}, for action localization in \sect{experiments-action-localization}, and for event recognition in \sect{experiments-event-recognition}.

\subsection{Action recognition}
\label{sec:experiments-action-recognition}

We first compare bag-of-words (BOW) and Fisher vectors (FV) for feature encoding, and evaluate the performance gain due to different motion stabilization steps.
Then, we assess the impact of  removing inconsistent matches based on human detection, and finally compare to the state of the art.

\subsubsection{Feature encoding with BOW and FV}
We begin our experiments with the original non-stabilized MBH descriptor~\citep{Wang2013} and compare its performance using BOW and FV under different parameter settings.
For this initial set of experiments, we chose the Hollywood2 and HMDB51 datasets as they are widely used and are representative in difficulty and size for the task of action recognition.
We evaluate the effect of including weak geometric information using the spatial Fisher vector (SFV) and spatio-temporal pyramids (STP).
We consider STP grids that divide the video in two temporal parts (T2), and/or three spatial horizontal parts (H3). 
When using STP, we always concatenate the representations (\ie, BOW or FV) over the whole video.
For the case of T2$+$H3, we concatenate all six BOW or FV representations (one for the whole video, two for T2, and three for H3).
Unlike STP, the SFV has only a limited effect for FV on the representation size, as it just adds six dimensions (for the spatio-temporal means and variances) for each visual word. 
For the BOW representation, the situation is different, since in that case there is only a single count per visual word, and the additional six dimensions of the SFV multiply the signature size by a factor seven; similar to the factor six for STP.

Table~\ref{tab:bov-vs-fv} lists all the results using different settings on Hollywood2 and HMDB51.
It is obvious that increasing the number of Gaussians $K$ leads to significant performance gain for both BOW and FV. However, the performance of FV tends to saturate after $K=256$, whereas BOW keeps improving up to $K=1024$. This is probably due to the high dimensionality of FV which results in an earlier saturation.
Both BOW and FV benefit from including STP and SFV, which are complementary since the best performance is always obtained when they are combined.

As expected, the RBF-$\chi^2$ kernel works better than the linear kernel for BOW. Typically, the difference is around 4-5\% on both Hollywood2 and HMDB51. When comparing different feature encoding strategies, the FV usually outperforms BOW by 6-7\% when using the same number of visual words. Note that FV of 64 visual words is even better than BOW of 1024 visual words; confirming that for FV fewer visual words are needed than for BOW. 

\setcounter{table}{1}
\begin{table*}  
    \centering
    \begin{tabular}{|l||cccc||cccc|}
      \hline
                          & \multicolumn{4}{c||}{Hollywood2}                 & \multicolumn{4}{c|}{HMDB51}                    \\
                          &      Baseline&      WarpFlow&       RmTrack&      Combined&      Baseline&      WarpFlow&       RmTrack&    Combined\\
      \hline\hline
          HOG             &        51.3\%&        52.1\%&        52.6\%&        53.0\%&        42.0\%&        43.1\%&        44.7\%&      44.4\%\\
          HOF             &        56.4\%&        61.5\%&        57.6\%&        62.4\%&        43.3\%&        51.7\%&        45.3\%&      52.3\%\\
          MBH             &        61.3\%&        63.1\%&        63.1\%&        63.6\%&        53.2\%&        55.3\%&        55.9\%&      56.9\%\\
      \hline
          HOG$+$HOF       &        61.9\%&        64.3\%&        63.2\%&        65.3\%&        51.9\%&        56.5\%&        54.2\%&      57.5\%\\
          HOG$+$MBH       &        63.0\%&        64.2\%&        63.6\%&        64.7\%&   {\bf56.3\%}&        57.8\%&        57.7\%&      58.7\%\\
          HOF$+$MBH       &        62.0\%&        65.3\%&        62.7\%&        65.2\%&        53.2\%&        57.1\%&        54.8\%&      58.3\%\\
      \hline
          HOG$+$HOF$+$MBH &   {\bf63.6\%}&   {\bf65.7\%}&   {\bf65.0\%}&   {\bf66.8\%}&        55.9\%&   {\bf59.6\%}&   {\bf57.8\%}& {\bf60.1\%}\\
      \hline

    \end{tabular}
    \caption{Comparison of baseline to our method and intermediate. WarpFlow: computing HOF and MBH using warped optical flow, while keeping all the trajectories. RmTrack: removing background trajectories, but compute descriptors using the original flow. Combined: removing background trajectories, and descriptors on warped flow. All the results use SFV$+$STP, $K=256$, and human detection to remove outlier matches.} 
    \label{tab:compare-all}
\end{table*}

We further explore the limits of BOW performance by using very large vocabularies, \ie, with $K$ up to $32,768$.
The results are shown in the left panel of Figure~\ref{fig:large_k}.
For BOW, we use $\chi^2$ kernel and T2$+$H3 which give the best results in Table~\ref{tab:bov-vs-fv}.
For a fair comparison, we only use T2$+$H3 for FV without SFV.
On both Hollywood2 and HMDB51, the performance of BOW becomes saturated when $K$ is larger than $8,192$.
If we compare BOW and FV representations with similar dimensions (\ie, $K=32,768$ for BOW and $K$ between $64$ and $128$ for FV), FV still outperforms BOW by $2\%$ on HMDB51 and both have comparable performance for Hollywood2.
Moreover, feature encoding with large vocabularies is very time-consuming as shown in the right panel of Figure~\ref{fig:large_k}, where $K=32,768$ for BOW is eight times slower than $K=128$ for FV.
This can impose huge computational cost for large datasets such as \trecvid MED.
FV is also advantageous as it achieves excellent results with a linear SVM which is more efficient than kernel SVMs.
Note however, that the classifier training time is negligible compared to the feature extraction and encoding time, \eg it only takes around 200 seconds for FV with $K=256$ to compute the Gram matrix and to train the classifiers on the Hollywood2 dataset.


To sum up, we choose FV with both STP and SFV, and set $K=256$ for a good compromise between accuracy and computational complexity. We use this setting in the rest of experiments unless stated otherwise.

\subsubsection{Evaluation of improved trajectory features}
We choose the dense trajectories~\citep{Wang2013} as our baseline, compute HOG, HOF and MBH descriptors as described in Section~\ref{sec:extract_dtf}, and report results on all the combinations of them.
In order to evaluate intermediate results, we decouple our method into two parts,
\ie, ``WarpFlow" and ``RmTrack", which stand for warping optical flow
with the homography and
removing background trajectories consistent with the homography. The combined setting uses both.
The results are presented in Table \ref{tab:compare-all} for Hollywood2 and HMDB51.

In the following, we discuss the results per descriptor.
The results of HOG are similar for different variants on
both datasets.
Since HOG is designed to capture static appearance information,
we do not expect that compensating camera motion significantly improves its performance.

HOF benefits the most from stabilizing optical flow. Both ``Combined" and ``WarpFlow" are substantially better than the other two. On Hollywood2, the improvements are around
$5\%$. On HMDB51,
the improvements are even higher: around $10\%$. After motion compensation, the performance of HOF
is comparable to that of MBH.

\setcounter{table}{2}
\begin{table*} [t] 
    \centering
    \begin{tabular}{|l||cccc||cccc|}
      \hline
                          &      \multicolumn{4}{c||}{Hollywood2-sub}                 &      \multicolumn{4}{c|}{High Five}                    \\
                          &      Baseline&           Non&     Automatic&        Manual&      Baseline&           Non&     Automatic&      Manual\\
      \hline\hline
          HOG             &        39.9\%&        40.0\%&        39.7\%&        40.4\%&        48.2\%&        49.7\%&        49.3\%&      50.2\%\\
          HOF             &        40.7\%&        49.6\%&        51.5\%&        52.1\%&        53.4\%&        66.8\%&        67.4\%&      68.1\%\\
          MBH             &        49.6\%&        52.5\%&        53.1\%&        54.2\%&        61.5\%&        67.3\%&        68.5\%&      68.8\%\\
      \hline
          HOG$+$HOF       &        46.3\%&        49.9\%&        51.3\%&        52.8\%&        57.5\%&        66.3\%&        67.5\%&      67.5\%\\
          HOG$+$MBH       &        49.8\%&        51.5\%&        52.3\%&        53.4\%&        61.8\%&        66.9\%&        67.2\%&      67.8\%\\
          HOF$+$MBH       &        49.6\%&        53.8\%&        54.4\%&        55.3\%&        61.4\%&   {\bf69.1\%}&   {\bf70.5\%}& {\bf71.2\%}\\
      \hline
          HOG$+$HOF$+$MBH &   {\bf50.8\%}&   {\bf54.3\%}&   {\bf55.5\%}&   {\bf56.3\%}&   {\bf62.5\%}&        68.1\%&        69.4\%&      69.8\%\\
      \hline
    \end{tabular}
    \caption{Impact of human detection on a subset of Hollywood2 and High Five datasets. ``Baseline": without motion stabilization; ``Non": without human detection; ``Automatic": automatic human detection; ``Manual": manually annotation. As before, we use SFV$+$STP, and set $K=256$.}
    \label{tab:human-detection}
\end{table*}

MBH is known for its robustness to camera motion~\citep{Wang2013}. However, its performance still improves,
as motion boundaries are much clearer, see \fig{compare_flow} and \fig{human_detect}.
We have over $2\%$ improvement on both datasets. 

Combining HOF and MBH further improves the results as they are complementary to each other.
HOF represents zero-order motion information, whereas MBH focuses on first-order derivatives.
Combining all three descriptors
achieve the best performance, as shown in the last row of Table \ref{tab:compare-all}.

\begin{table*} 
\centering \small
\begin{tabular}{|cc|cc|cc|}
  \hline
  \multicolumn{2}{|c|}{Hollywood2}                 & \multicolumn{2}{c|}{HMDB51}                           & \multicolumn{2}{c|}{Olympic Sports} \\
  \hline\hline
   \citealp{Jiang2012}           & 59.5\%                     & \citealp{Jiang2012}           & 40.7\%      & \citealp{Jain2013}       & 83.2\%      \\
   \citealp{Mathe2012}           & 61.0\%                     & \citealp{Ballas2013}          & 51.8\%      & \citealp{Li2013dynamic}  & 84.5\%      \\
   \citealp{Zhu2013}             & 61.4\%                     & \citealp{Jain2013}            & 52.1\%      & \citealp{Wang2013mining} & 84.9\%      \\
   \citealp{Jain2013}            & 62.5\%                     & \citealp{Zhu2013}             & 54.0\%      & \citealp{Gaidon2013}     & 85.0\%      \\
  \hline
   Baseline                       & 63.6\%                     & Baseline                       & 55.9\%      & Baseline                  & 85.8\%      \\
   Without HD                     & 66.1\%                     & Without HD                     & 59.3\%      & Without HD                & 89.6\%      \\
   With HD                        & {\bf66.8\%}                & With HD                        & {\bf60.1\%} & With HD                   & {\bf90.4\%}  \\
  \hline
  \hline
  \multicolumn{2}{|c|}{High Five} & \multicolumn{2}{c|}{UCF50} & \multicolumn{2}{c|}{UCF101} \\
  \hline\hline
  \citealp{Ma2013}               & 53.3\%                     & \citealp{Shi2013}             & 83.3\%      & \citealp{Peng2013hybrid} & 84.2\%  \\
  \citealp{Yu2012}               & 56.0\%                     & \citealp{Wang2013mining}      & 85.7\%      & \citealp{Murthy2013}     & 85.4\%  \\
  \citealp{Gaidon2013}           & 62.4\%                     & \citealp{Ballas2013}          & \bf 92.8\%  & \citealp{Karaman2013}    & 85.7\%  \\
  \hline
   Baseline                       & 62.5\%                     & Baseline                       & 89.1\%      & Baseline                  & 83.5\%      \\
   Without HD                     & 68.1\%                     & Without HD                     & 91.3\%      & Without HD                & 85.7\%      \\
   With HD                        & {\bf69.4\%}                & With HD                        & 91.7\%      & With HD                   & {\bf86.0\%} \\
  \hline
\end{tabular}
\caption{Comparison of our results (HOG$+$HOF$+$MBH) to the state of art. We present our
  results for FV encoding ($K=256$) using SFV$+$STP both with and without automatic human
  detection (HD). Best result for each dataset is marked in bold.}
\label{tab:stateofart}
\end{table*}

\subsubsection{Removing inconsistent matches due to humans}
We investigate the impact of removing inconsistent
matches due to humans when estimating
the homography, see Figure~\ref{fig:human_detect} for an illustration.
We compare four cases: (i) the baseline without stabilization,
(ii) estimating the homography without human detection, (iii) with automatic human detection, and (iv) with manual labeling of humans.
This allows us to measure the impact of removing matches from human
regions as well as to determine an upper bound in case of a perfect human
detector. We consider two datasets: Hollywood2 and High Five. To limit the labeling effort on Hollywood2, we annotated humans in 20
training and 20 testing videos for each action class. On High Five, we use the annotations provided by the authors of~\citep{Patron2010}.

As shown in Table \ref{tab:human-detection}, human detection helps to improve motion descriptors (\ie, HOF and MBH), since removing inconsistent matches on humans improves the homography estimation. Typically, the improvements are over $1\%$ when using an automatic human detector or manual labeling.
The last two rows of Table \ref{tab:stateofart} show the impact of automatic human detection
on all six datasets. Human detection always improves the performance slightly.

\subsubsection{Comparison to the state of the art}
Table \ref{tab:stateofart} compares our method with the most recent
results reported in the literature.
On Hollywood2, all presented
results~\citep{Jain2013,Jiang2012,Mathe2012,Zhu2013}
improve dense trajectories in different ways. 
\citet{Mathe2012}
prune background features based on visual saliency.
\citet{Zhu2013} apply multiple instance learning on top of dense
trajectory features in order to learn mid-level ``acton" to better represent human actions.
Recently, \citet{Jain2013} report $62.5\%$ by decomposing visual motion to stabilize dense trajectories.
We further improve their results by over $4\%$.

HMDB51~\citep{Kuehne2011} is a relatively new dataset.
\citet{Jiang2012} achieve $40.7\%$ by modeling the relationship between dense trajectory clusters.
\citet{Ballas2013} report $51.8\%$ by pooling dense trajectory features from
regions of interest using video structural cues estimated by different
saliency functions.
The best previous  result is from \citep{Zhu2013}. We improve it further by over $5\%$, and obtain $60.1\%$ accuracy.

Olympic Sports~\citep{Niebles2010} 
contains significant camera motion, which results in a large number of trajectories in the background.
\citet{Li2013dynamic} report $84.5\%$ by dynamically pooling feature from the most
informative segments of the video.
\citet{Wang2013mining} propose motion atom and phrase as a mid-level
temporal part for representing and classifying complex action, and achieve $84.9\%$.
\citet{Gaidon2013} model the motion hierarchies of dense trajectories \citep{Wang2013} with tree structures and report $85.0\%$.
Our improved trajectory
features outperform them by over $5\%$.

High Five~\citep{Patron2010} focuses on human interactions and serves as a good testbed for
various structure model applied for action recognition. \citet{Ma2013} propose hierarchical space-time
segments as a new representation for simultaneously action recognition and localization. They only extract
the MBH descriptor from each segment and report $53.3\%$ as the final performance. \citet{Yu2012} propagate Hough voting
of STIP~\citep{Laptev2008} features in order to overcome their sparseness, and achieve $56.0\%$.
With our framework we achieve $69.4\%$ on this challenging dataset.

\begin{table}
  \centering
  \begin{tabular}{|l|c||c|c|c|c|}
    \hline
      & \begin{sideways}Overlap      \end{sideways}
      & \begin{sideways}Drinking     \end{sideways}  
      & \begin{sideways}Smoking      \end{sideways}  
      & \begin{sideways}Open door \; \end{sideways}  
      & \begin{sideways}Sit Down     \end{sideways}  
      \\
    \hline\hline
      NMS      & 20 & 73.2\%     & 32.3\%     & 23.3\%     & 28.6\%     \\
    \hline                                                              
      RS-NMS   & 20 & 76.5\%     & 38.0\%     & 23.2\%     & 26.6\%     \\
      DP-NMS   & 0  & 71.4\%     & 36.7\%     & 21.0\%     & 23.6\%     \\
    \hline
      NMS      & 0  & 74.1\%     & 32.4\%     & 24.2\%     & \bf 28.9\% \\
      RS-NMS   & 0  & \bf 80.2\% & \bf 40.9\% & \bf 26.0\% & 27.1\%     \\
    \hline
  \end{tabular}
  \caption{%
    Evaluation of the non-maximum suppression variants:
    classic non-maximum suppression (NMS),
    dynamic programming non-maximum suppression (DP-NMS),
    and re-scored non-maximum suppression (RS-NMS).
    The overlap parameter (second column) indicates the maximum overlap (intersection over union) allowed between any two windows after non-maximum suppression.
    We use HOG$+$HOF$+$MBH from improved trajectory features (without human detector) with FV ($K=256$) augmented by SFV$+$STP.
  }
  \label{tab:nms-variants-stab}
\end{table}

\begin{table*}


    \centering
    \begin{tabular}{|l||ccc||ccc|}
       \hline
                          & \multicolumn{3}{c||}{Drinking}       & \multicolumn{3}{c|}{Smoking} \\
                          & Baseline & Without HD & With HD & Baseline & Without HD & With HD   \\
      \hline\hline
          HOG             & 44.3\%   & 52.7\%     & 51.5\%  & 31.0\%   & 32.9\%     & 33.9\% \\
          HOF             & 82.5\%   & 79.2\%     & 79.1\%  & 28.9\%   & 34.7\%     & 33.9\% \\
          MBH             & 78.7\%   & 73.0\%     & 70.4\%  &\bf47.7\% &\bf48.7\%   & 43.2\% \\
      \hline
          HOG$+$HOF       & 80.8\%   &\bf81.1\%   &\bf79.9\%& 35.5\%   & 33.5\%     & 33.0\% \\
          HOG$+$MBH       & 78.2\%   & 74.3\%     & 75.0\%  & 40.5\%   & 42.7\%     & 42.3\% \\
          HOF$+$MBH       &\bf85.0\% & 79.0\%     & 78.3\%  & 46.8\%   & 45.7\%     &\bf45.0\% \\
      \hline
          HOG$+$HOF$+$MBH & 81.6\%   & 80.2\%     & 79.0\%  & 38.5\%   & 40.9\%     & 39.4\% \\
      \hline\hline
                          & \multicolumn{3}{c||}{Open door}      & \multicolumn{3}{c|}{Sit down}        \\
                          & Baseline   & Without HD & With HD    & Baseline   & Without HD & With HD    \\
      \hline\hline
          HOG             & 21.6\%     & 23.8\%     & 21.4\%     & 14.9\%     & 14.3\%     & 14.3\% \\
          HOF             & 21.4\%     & 19.8\%     & 23.9\%     & 25.5\%     & 25.5\%     & 23.8\% \\
          MBH             & 29.5\%     & 23.4\%     & 22.9\%     & 26.1\%     & 25.8\%     & 25.6\% \\
      \hline                                                                                          
          HOG$+$HOF       & 20.9\%     & 27.5\%     & 26.9\%     & 24.1\%     & 21.9\%     & 22.6\% \\
          HOG$+$MBH       &\bf29.6\%   &\bf30.2\%   &\bf29.2\%   & 28.3\%     & 25.0\%     & 25.2\% \\
          HOF$+$MBH       & 28.8\%     & 23.4\%     & 23.8\%     &\bf30.6\%   &\bf27.2\%   & 27.1\% \\
      \hline                                                                                          
          HOG$+$HOF$+$MBH & 28.8\%     & 26.0\%     & 26.4\%     & 29.6\%     & 27.1\%     &\bf27.6\% \\
      \hline
    \end{tabular}
    \caption{%
    Comparison of improved trajectory features (with and without human detection) to the baseline for the action localization task.
    We use Fisher vector ($K=256$) with SFV$+$STP to encode local descriptors, and apply RS-NMS-0 for non-maxima suppression.
    We show results on two datasets: the \emph{Coffee \& Cigarettes} dataset~\citep{Laptev2007} (\emph{drinking} and \emph{smoking}) and the DLSBP dataset~\citep{Duchenne2009} (\emph{Open Door} and \emph{Sit Down}).%
    }
    \label{tab:localization-compare-all}
\end{table*}

UCF50~\citep{Reddy2012} can be considered as an extension of the widely used YouTube dataset~\citep{Liu2009}.
Recently, \citet{Shi2013} report $83.3\%$ using randomly sampled HOG, HOF, HOG3D and MBH descriptors.
\citet{Wang2013mining} achieve $85.7\%$. The best result so far is $92.8\%$ from \citet{Ballas2013}.
We obtain a similar accuracy of $91.7\%$.

UCF101~\citep{Soomro2012} is used in the recent THUMOS'13 Action Recognition Challenge~\citep{THUMOS13}.
All the top results are built on different variants of dense trajectory features~\citep{Wang2013}.
\citet{Karaman2013} extract many features (such as HOG, HOF, MBH, STIP, SIFT, \etc) and do late fusion with logistic regression to combine the output of each feature channel. \citet{Murthy2013} combine ordered trajectories~\citep{Murthy2013ordered} and improved trajectories \citep{Wang2013a}, and apply Fisher vector to encode them.
With our framework we obtained $86.0\%$, and ranked first among all 16 participants.

\subsection{Action localization}
\label{sec:experiments-action-localization}

In our second set of experiments we consider the localization of four actions (\ie, \emph{drinking}, \emph{smoking}, \emph{open door} and \emph{sit down}) in feature length movies.
We set the encoding parameters the same as action recognition: $K=256$ for Fisher vector with SFV$+$STP.
We first consider the effect of different NMS variants using our improved trajectory features without human detection.
We then compare with the baseline dense trajectory features and discuss the impact of human detection.
Finally we present a comparison to the state-of-the-art methods.

\subsubsection{Evalution of NMS variants}

We report all the results by combining HOF, HOF and MBH together, and present them in Table~\ref{tab:nms-variants-stab}. 
We see that simple rescoring (RS-NMS) significantly improves over standard NMS on two out of four classes, while the dynamic programming version (DP-NMS) is slightly inferior when compared with RS-NMS.
To test whether this is due to the fact that DP-NMS does not allow any overlap, we also test  NMS and RS-NMS with zero overlap.
The results show that for standard NMS zero or 20\% overlap does not significantly change the results on all four action classes, while for RS-NMS zero overlap is beneficial on all classes.
Since RS-NMS zero overlap performs the best among all five different variants, we use it in the remainder of the experiments.

\subsubsection{Evaluation of improved trajectory features}


We present detailed experimental results in Table~\ref{tab:localization-compare-all}. We analyze all the combinations of the three descriptors and compare our improved trajectory features (with and without human detection) with the baseline dense trajectory features.

We observe that combining all  descriptors usually gives better performance than individual descriptors.
The improved trajectory features are outperformed by the baseline on three out of four classes for the case of HOG$+$HOF$+$MBH.
Note that the results of different descriptors and settings are less consistent than they are on action recognition datasets, \eg, Table~\ref{tab:compare-all}, as here we report the results for each class separately. Furthermore, since the action localization datasets are much smaller than action recognition ones, the number of positive examples per category is limited, which renders the experimental results less stable. In randomised experiments, where we leave one random positive test sample out from the test set, we observe standard deviations of the same order as the differences between the various settings (not shown for sake of brevity). 

As for the impact of human detection, surprisingly leaving it out  performs better for \emph{drinking} and \emph{smoking}.
Since \emph{Coffee \& Cigarettes} essentially consists of scenes with static camera, this result might be due to inaccuracies in the homography estimation. 



\begin{table}
  \centering
  \begin{tabular}{|l||c|c|c|c|}
  \hline
    & \begin{sideways}Drinking\end{sideways}
    & \begin{sideways}Smoking \end{sideways}
    & \begin{sideways}Open door \; \end{sideways}
    & \begin{sideways}Sit Down \end{sideways}
    \\
  \hline\hline
    \citealp{Laptev2007}                                              & 49.0\%     & ---        & ---        & ---    \\
    \citealp{Duchenne2009}                                            & 40.0\%     & ---        & 14.4\%     & 13.9\% \\
    \citealp{Klaser2010b}                                             & 54.1\%     & 24.5\%     & ---        & ---    \\
    \citealp{Gaidon2011}                                              & 57.0\%     & 31.0\%     & 16.4\%     & 19.8\% \\
  \hline
    RS-NMS zero overlap                                             & \bf 80.2\% & \bf 40.9\% & \bf 26.0\% & \bf 27.1\% \\
  \hline
  \end{tabular}
  \caption{%
  Improved trajectory features  without human detection compared to the state of the art for localization.
  We use HOG$+$HOF$+$MBH descriptors encoded with FV ($K=256$) and SFV$+$STP, and apply RS-NMS zero overlap for non-maxima suppression. 
  }
  \label{tab:state-of-the-art-localization}
\end{table}

\begin{table*}
  \begin{center}
  {\footnotesize
  \begin{tabular}{|l||cccccccccc|c|}
      \hline
        &
        \begin{sideways}Birthday party\end{sideways} &
        \begin{sideways}Changing a vehicle tire\end{sideways}  &
        \begin{sideways}Flash mob gathering\end{sideways}  &
        \begin{sideways}Getting vehicle unstuck\;\end{sideways}  &
        \begin{sideways}Grooming an animal\end{sideways}  &
        \begin{sideways}Making a sandwich\end{sideways}  &
        \begin{sideways}Parade\end{sideways}  &
        \begin{sideways}Parkour\end{sideways}  &
        \begin{sideways}Repairing an appliance\end{sideways}  &
        \begin{sideways}Sewing project\end{sideways}  &
        Mean \\
      \hline\hline
        HOG                   & 28.7\%    & 45.9\%    & 57.2\%    & 38.6\%    & 18.5\%    & 21.1\%    & 41.4\%    & 51.5\%    & 41.1\%    & 25.8\%    & 37.0\%    \\ 
        HOF                   & 18.8\%    & 28.5\%    & 54.6\%    & 37.2\%    & 24.5\%    & 17.2\%    & 44.9\%    & 66.7\%    & 35.6\%    & 28.5\%    & 35.7\%    \\ 
        MBH                   & 26.2\%    & 39.1\%    & 59.8\%    & 37.7\%    & 30.4\%    & 19.7\%    & 46.4\%    & 72.6\%    & 33.6\%    & 32.8\%    & 39.8\%    \\ 
      \hline
        HOG$+$HOF             & 27.6\%    & 49.9\%    & 59.8\%    & 45.1\%    & 30.6\%    & 22.4\%    & 48.4\%    & 69.4\%    & 40.8\%    & 35.0\%    & 42.9\%    \\ 
        HOG$+$MBH             & 30.8\%    & \bf53.9\% & 61.5\%    & 40.0\%    & 38.2\%    & \bf28.8\% & \bf53.4\% & 72.0\%    & 38.1\%    & \bf43.3\% & \bf46.0\% \\ 
        HOF$+$MBH             & 26.8\%    & 40.7\%    & 59.8\%    & 41.2\%    & 31.2\%    & 20.3\%    & 47.6\%    & 71.8\%    & 33.5\%    & 34.7\%    & 40.8\%    \\ 
      \hline
        HOG$+$HOF$+$MBH       & \bf31.3\% & 53.0\%    & \bf61.9\% & \bf47.4\% & \bf38.2\% & 23.4\%    & 51.4\%    & \bf73.2\% & \bf41.6\% & 37.5\%    & 45.9\%    \\ 
      \hline
  \end{tabular}
  }
  \end{center}
  \caption{Performance in terms of AP on the full \trecvid MED 2011 dataset.
  We use ITF and encode them with FV ($K=256$).
  We also use SFV and STP, but only with a horizontal stride (H3), and no temporal split (T2). We rescale the video to a maximal width of 
480 pixels.}
\label{tab:trecvid-full}
\end{table*}

\subsubsection{Comparison to the state of the art}

In \tab{state-of-the-art-localization}, we compare our RS-NMS zero overlap method with previously reported state-of-the-art results.
As features we use HOG$+$HOF$+$MBH of the improved trajectory features, but without human detection.
We obtain substantial improvements on all four action classes, 
despite the fact that previous work used more elaborate techniques.
For example, \citet{Klaser2010b} relied on human detection and tracking, while \citet{Gaidon2011} requires finer annotations that indicate the position of characteristic moments of the actions (actoms).
The biggest difference comes from the \emph{drinking} class, where our result is over 23\% better than that of \citet{Gaidon2011}.

\subsection{Event recognition}
\label{sec:experiments-event-recognition}

In our last set of experiments we consider the large-scale \trecvid MED 2011 event recognition dataset.
For this set of experiments, we do not use the human detector during homography estimation.
We took this decision for practical reasons:
running the human detector on $1,000$ hours of video would have taken more than two weeks on $500$ cores;
the speed is about $10$ to $15$ seconds per frame on a single core.
We also leave out the T2 split of STP, because of both performance and computational reasons.
We have found on a subset of \trecvid 2011 train data that the T2 of STP does not improve the results.
This happens because the events do not have a temporal structure that can be easily captured by the rigid STP, as opposed to the actions that are temporally well cropped.

\subsubsection{Evaluation of improved trajectory features}

\tab{trecvid-full} shows results on the \trecvid MED 2011 dataset.
We contrast the different descriptors and their combinations for all the ten event categories.
We observe that the MBH descriptors are best performing among the individual channels.
The fact that HOG outperforms HOF demonstrates that there is rich contextual appearance information in the scene as \trecvid MED contains complex event videos.

Between the two-channel combinations, the best one is HOG$+$MBH, followed by HOG$+$HOF and HOF$+$MBH.
This order is given by the complementarity of the features:
both HOF and MBH encode motion information, while HOG captures texture information.
Combining all three channels performs similarly to the best two-channel variant.

If we remove all spatio-temporal information (H3 and SFV), performance drops from %
$45.9$
to %
$43.8$.
This underlines the importance of weak geometric information, even for the highly unstructured videos found in  \trecvid MED.

We consider the effect of re-scaling the videos to different resolutions in \tab{tv11-resolution} for both baseline DTF and our ITF.
From the results we see that ITF always improves over DTF: even on low resolutions there are enough feature matches in order to estimate the homography reliably.
The performance of both DTF and ITF does not improve much when using higher resolutions than 320.

The results in \tab{tv11-resolution} also show that the gain from ITF on \trecvid MED is less pronounced than the gain observed for action recognition.
This is possibly due to the generally poorer quality of the videos in this dataset, \eg due to motion blur in videos recorded by hand-held cameras.
In addition, a major challenge in this data set is that for many videos the information characteristic for the category is limited to a relatively short sub-sequence of the video.
As a result the video representations are affected by background clutter from  irrelevant portions of the video.
This difficulty might limit the beneficial effects of our improved  features.

\tab{tv11-time} provides the speed of computing our video representations when using the settings from \tab{tv11-resolution}. 
    Computing ITF instead of DTF features increases the runtime by around of a factor of two.
    For our final setting (videos resized to 480 px width, improved dense trajectories, HOG, HOF, MBH, stabilized without the human detector and encoded with FV and H3 SPM and SFV),
    the slowdown factor with respect to the real video time is around $10\times$ on a single core.
    This translates in less than a day of computation for the 1,000 hours of TRECVID test data on a 500-core cluster.
    

\begin{table}
  \centering
  \begin{tabular}{|l||c|c|c|c|}
    \hline
     AP & $160$ px & $320$ px     & $480$ px & $640$ px \\
    \hline
     DTF & 40.6\% & 44.9\% & 43.0\% & 44.3\% \\
     ITF & 41.0\% & 45.6\% &\bf45.9\% & 45.4\% \\
    \hline
  \end{tabular}
  \caption{Comparison of our improved trajectory features (ITF) with the baseline dense trajectory features (DTF) for different resolutions
  on the \trecvid MED dataset. For both ITF and DTF, we combine HOG, HOF and MBH, and use FV ($K=256$) augmented with SFV  and STP, but only use H3 and not T2 for STP.}
  \label{tab:tv11-resolution}
\end{table}

\begin{table}
  \centering
  \begin{tabular}{|l||cc|cc|cc|cc|}
    \hline
     FPS & \multicolumn{2}{c|}{$160$ px} & \multicolumn{2}{c|}{$320$ px}     & \multicolumn{2}{c|}{$480$ px} & \multicolumn{2}{c|}{$640$ px} \\
    \hline
     DTF &  40.8&83.4  &  10.4&22.1  &  4.5&9.2   &  2.1&5.2 \\
     ITF &  18.5&91.7  &   5.1&23.8  &  2.2&10.2  &  1.2&5.9 \\
    \hline
  \end{tabular}
  \caption{The speed (frames per second) of computing our proposed video representation using different resolutions on the \textsc{TRECVID} MED dataset; left: the speed of computing raw features (\ie, DTF or ITF); right: the speed of encoding the features into a high dimensional Fisher vector ($K=256$).}
  \label{tab:tv11-time}
\end{table}

\begin{table}
  \begin{center}
  {\footnotesize
  \begin{tabular}{|l|l|r|}
      \hline
        Paper & Features & mAP \\
      \hline\hline
        \citealp{Tang2012}      & HOG3D                            & $4.8\%$ \\
        \citealp{Vahdat2013a}   & HOG3D, textual information       & $8.4\%$ \\
        \citealp{Kim2013}       & HOG3D, MFCC                      & $9.7\%$ \\
        \citealp{Li2013dynamic} & STIP                             & $12.3\%$ \\
        \citealp{Vahdat2013b}   & HOG3D, SSIM, color,              & {$15.7\%$} \\
                                & sparse and dense SIFT            & \\
        \citealp{Tang2013}      & HOG3D, ISA, GIST, HOG,           & {$21.8\%$} \\
                                & SIFT, LBP, texture, color        & \\
      \hline
        ITF                     & HOG, HOF, MBH                    & $\bf31.6\%$ \\
      \hline
  \end{tabular}
  }
  \end{center}
  \caption{Performance in terms of AP on the \trecvid MED 2011 dataset using the EVENTS/DEV-O split.
  The feature settings are the same as Table~\ref{tab:trecvid-full}:
  improved trajectory features (HOG$+$HOF$+$MBH), encoded with FV ($K=256$) and SFV$+$H3.
  }
\label{tab:trecvid-dev-o-related-work}

\end{table}

\subsubsection{Comparison to the state of the art}

We compare to the state-of-the-art in \tab{trecvid-dev-o-related-work}.
We consider the EVENTS/DEV-O split of the \trecvid MED 2011 dataset,
since most results are reported using this setup.

The top three results were reported by the following authors.
\citet{Li2013dynamic} attained  12.3\% by automatically segmenting videos into coherent sub-sequences over which the features are pooled.
\citet{Vahdat2013b} achieved 15.7\%  by  using multiple kernel learning to combine different features, and latent variables to infer the relevant portions of the videos.
\citep{Tang2013} obtained the best reported result so far of 21.8\%, using a method based on AND-OR graphs to combine a large set of features in different subsets.

We observe a dramatic improvement when comparing our result of 31.6\% to the state of the art.
In contrast to these other approaches, our work focuses on good local features and their encoding, and then learns a linear SVM classifier over concatenated Fisher vectors computed from the HOG, HOF and MBH descriptors.

\section{Conclusions}
\label{sec:conclusion}

This paper improves dense trajectories by explicitly estimating camera motion.
We show that the performance can be significantly improved by removing background
trajectories and warping optical flow with a robustly estimated
homography approximating the camera motion. Using a state-of-the-art human detector,
possible inconsistent matches can be removed during camera motion estimation,
which makes it more robust. We also explore Fisher vector as an alternative feature
encoding approach to bag-of-words histograms, and consider the effect of spatio-temporal pyramids and spatial Fisher vectors to encode weak geometric layouts.

 An extensive evaluation on
three challenging tasks ---action recognition, action localization in movies, and  complex event recognition---
demonstrates the effectiveness and flexibility of our new framework.
We also found that action localization results can be substantially improved by using a simple re-scoring technique before applying NMS, to suppress a bias for too  short windows.
Our proposed pipeline significantly outperform the state of the art on all three tasks.
Our approach can serve as a general pipeline for various video recognition problems.

\vspace{2mm} \noindent {\bf Acknowledgments.} This work was supported
by Quaero (funded by OSEO, French State agency for innovation),  the
European integrated project AXES, the MSR/INRIA joint project and the
ERC advanced grant ALLEGRO.


\small
\bibliographystyle{spbasic}       


\end{document}